\documentclass{article}

\usepackage[main, preprint]{neurips_2026}
\makeatletter
\renewcommand{\@noticestring}{\phantom{Preprint. Work in progress.}}
\makeatother

\usepackage[utf8]{inputenc}
\usepackage[T1]{fontenc}

\usepackage{hyperref}
\usepackage{url}
\usepackage{booktabs}
\usepackage{amsfonts}
\usepackage{amssymb}
\usepackage{amsmath}
\usepackage{nicefrac}
\usepackage{microtype}
\usepackage[table]{xcolor}
\usepackage{colortbl}
\usepackage{graphicx}
\usepackage{subcaption}
\usepackage{capt-of}
\usepackage{multirow}
\usepackage{enumitem}
\usepackage{makecell}
\usepackage{array}
\usepackage{tabularx}
\usepackage{adjustbox}
\usepackage{float}
\usepackage{placeins}
\usepackage{arydshln}
\usepackage{etoolbox}

\definecolor{rateblue}{RGB}{56,96,170}
\definecolor{linkpink}{RGB}{255,20,147}
\definecolor{lightblueblock}{RGB}{240,248,255}
\definecolor{midblueblock}{RGB}{230,240,255}
\definecolor{deepblueblock}{RGB}{220,230,255}
\definecolor{lightbeigeblock}{RGB}{255,250,230}
\definecolor{lightgrayblock}{RGB}{232,232,232}
\newcommand{\ratecell}[2]{#1~{\textcolor{rateblue}{\scriptsize(#2)}}}
\makeatletter
\newcommand{\citeauthornolink}[1]{%
  \begingroup
  \let\hyper@natlinkstart\@gobble
  \let\hyper@natlinkend\relax
  \citeauthor{#1}%
  \endgroup
}
\newcommand{\methodvenue}[3]{#1~(\hyper@natlinkstart{#3}\citeauthornolink{#3}, \textit{#2}\hyper@natlinkend)}
\makeatother
\hypersetup{
  colorlinks=true,
  allcolors=linkpink,
  hypertexnames=false
}
\makeatletter
\patchcmd{\@maketitle}
  {\begin{tabular}[t]{c}\bf\rule{\z@}{24\p@}\@author\end{tabular}}
  {\begin{tabular}[t]{c}\bf\rule{\z@}{8\p@}\@author\end{tabular}}{}{}
\patchcmd{\@maketitle}{\vskip 0.3in \@minus 0.1in}{\vskip 0pt}{}{}
\patchcmd{\@toptitlebar}{\vskip 0.25in}{\vskip 0.12in}{}{}
\patchcmd{\@bottomtitlebar}{\vskip 0.29in}{\vskip 0.16in}{}{}
\patchcmd{\@bottomtitlebar}{\vskip 0.09in}{\vskip 2pt}{}{}
\makeatother

\title{Beyond Appearance Shifts: Task-Semantic Action Calibration for VLA Models}

\author{%
  {\bfseries Shuaijun Liu$^{1}$ \quad
  Feiyang You$^{1}$ \quad
  Chengyu Wu$^{1}$ \quad
  Shuyang Hao$^{1}$} \\[0pt]
  {\bfseries Chenglong Zhang$^{1}$ \quad
  Jingyao Cai$^{2}$ \quad
  Xingwei Chen$^{3,4}$ \quad
  Li Sun$^{5}$ \quad
  Ningxin Su$^{1}$\thanks{Corresponding author.}} \\[2pt]
  {\normalsize $^{1}$HKUST (Guangzhou) \quad
  $^{2}$NCCA, Bournemouth University \quad
  $^{3}$SJTU \enspace $^{4}$EIT, Ningbo \enspace $^{5}$BUPT} \\[1pt]
  {\normalsize \texttt{ningxinsu@hkust-gz.edu.cn} \quad
  Project Page: \url{https://nebulis-lab.com/BAS-VLA}}
}

\begin{document}

\maketitle
\noindent\begin{minipage}{\textwidth}
\centering
\makebox[\linewidth][c]{\includegraphics[width=\textwidth]{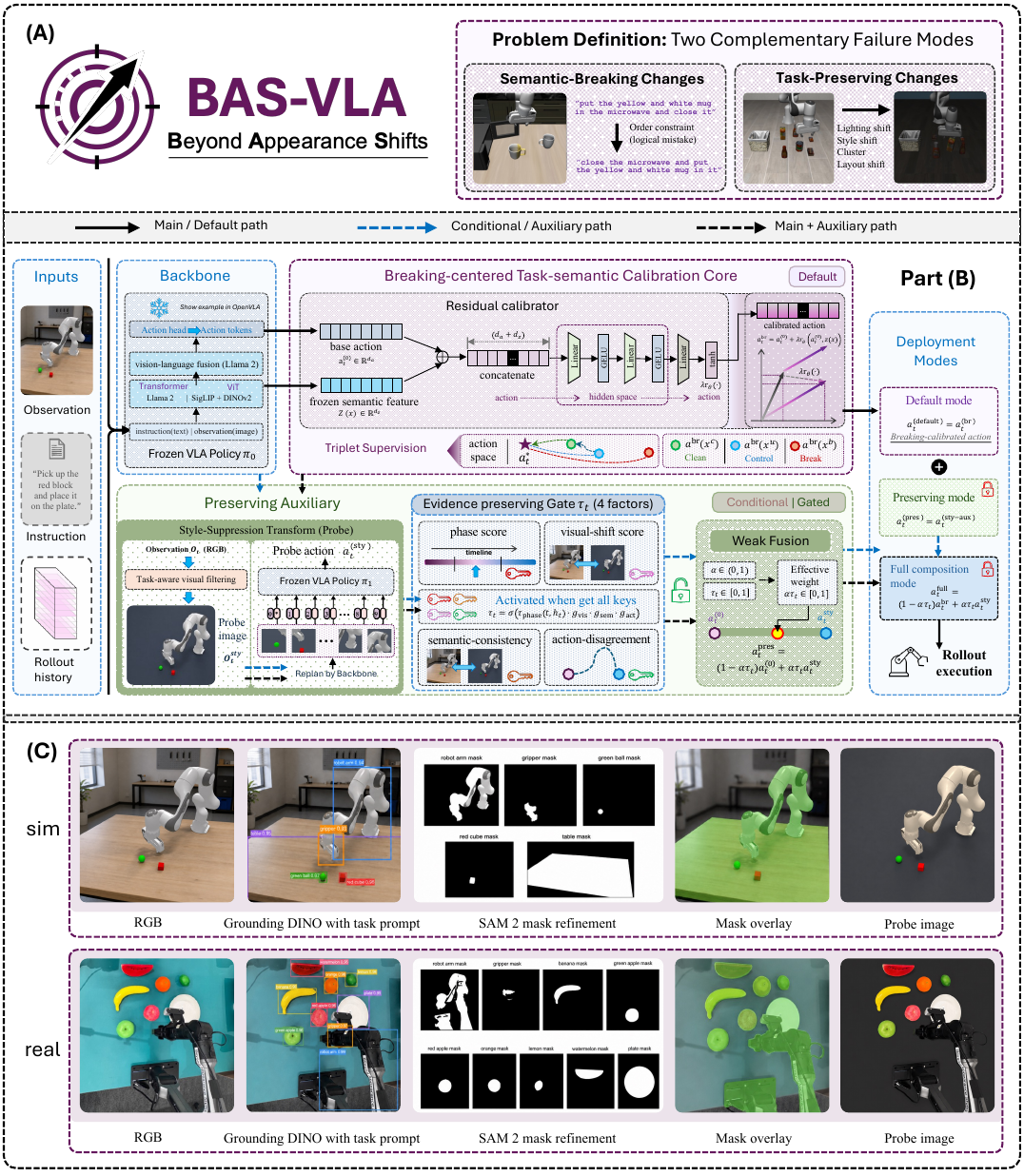}}
\parbox{\linewidth}{\captionof{figure}{BAS-VLA Overview. Part~(A) summarizes the two intervention types studied in this work: task-preserving changes and semantic-breaking changes. Part~(B) shows BAS-VLA as an action-calibration layer over a frozen base VLA. Part~(C) illustrates the task-aware visual filtering probe, from the input RGB image to text-conditioned grounding, SAM~2 mask refinement, merged task-relevant masks, and the resulting probe image $\tilde{o}_t^{\mathrm{sty}}$. \label{fig:overview}}}
\par
\end{minipage}
\clearpage
\begin{abstract}
Vision-language-action (VLA) models have achieved strong performance in embodied manipulation, but still struggle to distinguish between changes that should preserve behavior and those that require different actions. We identify two complementary failure modes. Under task-preserving changes, where task semantics remain unchanged but scene appearance varies (e.g., style, illumination, clutter, or paraphrasing), policies often exhibit unnecessary action drift. Conversely, under semantic-breaking changes, where key task semantics such as the target object or constraint are altered, policies frequently continue following the original trajectory instead of adapting to the new task. To address these failures, we propose BAS-VLA, a task-semantic action calibration framework built on top of a frozen base VLA. BAS-VLA adopts a breaking-centered calibration core as its default path, complemented by a selective evidence-gated preserving auxiliary that corrects appearance-induced drift when task-preserving evidence is strong. Across four target-object swap tasks, BAS-VLA improves changed-task success from 2.3\% to 68.5\% while maintaining strong clean and semantics-preserving performance and suppressing outdated behavior. On validated style-preserving shifts, it further improves success from 42\% to 70\% without degrading clean performance. These results demonstrate the importance of calibrating VLA policies to remain stable when task semantics are preserved and adapt when task semantics change.

\end{abstract}


\section{Introduction}

\begin{table}[!b]
\centering
\footnotesize
\setlength{\tabcolsep}{2pt}
\caption{Motivation evidence for reliability gaps before BAS-VLA. \emph{Original} denotes native carrier performance on the unmodified task condition, and \emph{Perturbed} denotes performance under the corresponding preserving or task-changed intervention. Carrier and suite names follow OpenPI-pi0.5~\citep{black2025pi05}, OpenVLA-OFT~\citep{kim2025oft}, and LIBERO~\citep{liu2023libero}. Here OpenVLA-OFT denotes an OpenVLA carrier built from the OFT fine-tuning recipe of~\citet{kim2025oft}; in our study, we adopt that recipe and perform our own fine-tuning for the simulation and real-robot settings used in the paper.}
\label{tab:motivation-gaps}
\vspace{0.15cm}
\begin{tabular}{>{\raggedright\arraybackslash}p{0.15\columnwidth}>{\raggedright\arraybackslash}p{0.17\columnwidth}>{\raggedright\arraybackslash}p{0.175\columnwidth}>{\centering\arraybackslash}p{0.175\columnwidth}>{\centering\arraybackslash}p{0.165\columnwidth}>{\centering\arraybackslash}p{0.10\columnwidth}}
\toprule
\textbf{Carrier} & \textbf{Suite} & \textbf{Perturbation} & \textbf{Original success} $\uparrow$ & \textbf{Perturbed success} $\uparrow$ & $\Delta$ (pts) \\
\midrule
OpenVLA-OFT & LIBERO-Spatial & Perimeter clutter & \cellcolor{lightblueblock}\ratecell{98/100}{98.0\%} & \cellcolor{lightblueblock}\ratecell{65/100}{65.0\%} & \cellcolor{midblueblock}$-33.0$ \\
OpenVLA-OFT & LIBERO-Spatial & Margin noise & \cellcolor{lightblueblock}\ratecell{97/100}{97.0\%} & \cellcolor{lightblueblock}\ratecell{51/100}{51.0\%} & \cellcolor{midblueblock}$-46.0$ \\
OpenPI-pi0.5 & LIBERO-Object & Illum. change & \cellcolor{lightblueblock}\ratecell{98/100}{98.0\%} & \cellcolor{lightblueblock}\ratecell{89/100}{89.0\%} & \cellcolor{midblueblock}$-9.0$ \\
OpenPI-pi0.5 & LIBERO-Object & Target-object Swap  & \cellcolor{lightblueblock}\ratecell{197/200}{98.5\%} & \cellcolor{lightblueblock}\ratecell{68/200}{34.0\%} & \cellcolor{midblueblock}$-64.5$ \\
\bottomrule
\end{tabular}
\end{table}

Vision-language-action (VLA) models have substantially broadened the scope of embodied manipulation by coupling language conditioning with large visuomotor policies~\citep{rt2,kim2024openvla}. Yet stronger backbones do not automatically yield \emph{reliable} behavior. A policy may solve a canonical instruction in a canonical scene, but still fail to preserve the right action when semantics are unchanged, or fail to revise the action when semantics genuinely change. This reliability gap is the focus of this work.

We study this gap through two intervention types. \emph{Task-preserving changes} alter style, illumination, clutter, viewpoint, or wording while leaving task semantics unchanged, so the desired behavior should remain in the same action basin. \emph{Semantic-breaking changes} modify a key semantic slot or constraint, so the behavior should separate decisively from the clean trajectory. Current VLAs can fail in both directions: they drift when they should stay stable, and they remain overly similar when they should change.

This problem touches several neighboring lines of work. Embodied grounding and affordance-aware control have improved how instructions connect to objects, skills, and constraints~\citep{saycan,mees2023affordances,huang2024rekep}, while counterfactual reasoning, semantic-intervention benchmarks, and process supervision have emphasized that correct outputs alone are insufficient when meaning changes~\citep{huyuk2025reasoning,pona2025abstract,chen2025counterbench,lightman2024lets,setlur2025rewarding,li2025pqm}. Our contribution is an explicit, asymmetric action-calibration objective for a frozen VLA: which edits should leave behavior invariant, which should force separation, and how should a shared policy be calibrated under both? This complements robustness benchmarks and backbone adaptation by explicitly calibrating action responses to semantics-preserving and semantics-changing interventions.

To address this gap, we propose \textbf{BAS-VLA}, a task-semantic action calibration framework built on top of a frozen base VLA. BAS-VLA does not replace the backbone or retrain a new foundation policy; it attaches a lightweight calibration layer to the action output of the shared VLA. Its default deployment path is a \emph{breaking-centered} calibration core that keeps semantically equivalent variants close while pushing genuine semantic breaks away from the clean action basin. In addition, BAS-VLA includes a selective \emph{evidence-gated preserving auxiliary} that activates only when nuisance evidence is strong while task semantics remain consistent, so that a weak stabilizing correction is beneficial.

\begin{figure}[tb]
\centering
\includegraphics[width=\linewidth]{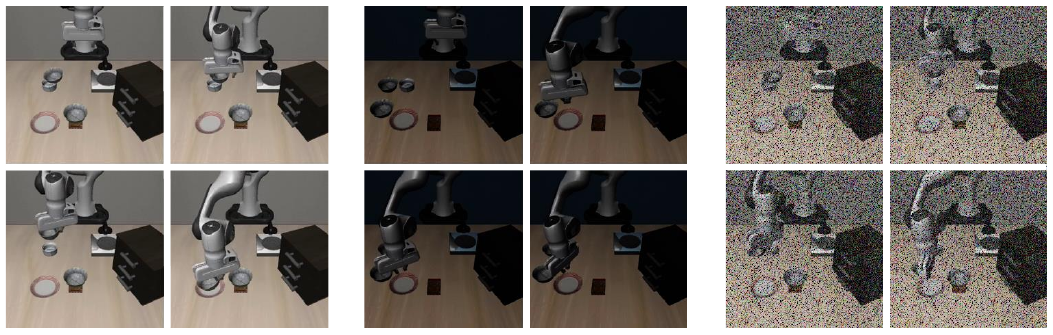}
\caption{Task-preserving motivation on a single Black-Bowl-to-Plate instruction. The three panels show a clean success, a failure under a darkened illumination shift, and a failure under a semantics-preserving noise perturbation.}
\vspace{-0.2cm}
\label{fig:motivation-triptych}
\end{figure}

Figure~\ref{fig:motivation-triptych} shows the qualitative preserving-side failure pattern, and Table~\ref{tab:motivation-gaps} quantifies the corresponding pre-method gap before BAS-VLA calibration. We evaluate BAS-VLA under a unified intervention protocol with matched backbones, task families, and rollout budgets. On the primary OpenPI-pi0.5 / LIBERO-Object Milk-Swap evaluation, BAS-VLA maintains 98.0\% clean success and 97.5\% control success while reducing clean-task success under the break condition to 0.0\%. On the Bowl-on-Ramekin style-shift setting, it raises changed-condition success from 42.0\% to 70.0\% without degrading the matched-clean reference row. In summary, we formulate VLA reliability as task-semantic action calibration, introduce BAS-VLA as a breaking-centered core with a selective evidence-gated preserving auxiliary, and provide matched-carrier evidence that reliable VLA behavior requires explicit task-semantic calibration rather than generic appearance robustness alone.
\vspace{-0.2cm}

\section{Related Work}
\vspace{-0.2cm}
Large-scale VLA systems provide the substrate on which BAS-VLA operates, but they are not the main intellectual neighbor. Generalist robot policies and multimodal foundation models have shown that broad pretraining improves language-conditioned manipulation across tasks and scenes~\citep{palme,rt2,octo_2023,kim2024openvla}. The question here is narrower: once a shared VLA is fixed, how should its action behavior respond to controlled semantic interventions?
The closest embodied line concerns language grounding, semantic control, and intervention-aware adaptation. SayCan couples language with affordance-weighted skill selection~\citep{saycan}, while later approaches ground language through visual affordances or explicit relational constraints~\citep{mees2023affordances,huang2024rekep}. Counterfactual feedback, semantic-intervention benchmarks, prompt-miscalibration analyses, and inference-time VLA adaptation methods all reinforce the same high-level point: correctness alone is too weak without sensitivity to meaning-preserving versus meaning-changing alternatives~\citep{huyuk2025reasoning,pona2025abstract,chen2025counterbench,cox2025mapping,black2025rtc,so2025sgac,liang2026aac,st4vla2026,wu2026pcd,koo2026hamlet,jang2026mgselect,lightman2024lets,setlur2025rewarding,li2025pqm}. BAS-VLA instead targets embodied \emph{action} calibration under matched manipulator rollouts. Appendix~\ref{app:extended-related} gives a compact extended positioning note.
LIBERO-Plus and LIBERO-PRO evaluate robustness and generalization under visual, language, object, and task changes~\citep{fei2025liberoplus,zhou2025liberopro}. Evaluation on these external protocols further tests BAS-VLA beyond the intervention settings used in our primary experiments.
\vspace{-0.2cm}

\section{Problem Setup}
\label{sec:problem-setup}
\vspace{-0.1cm}

\definecolor{badgeblue}{RGB}{255,228,228}
\definecolor{badgegreen}{RGB}{226,244,229}
\definecolor{badgeamber}{RGB}{238,229,255}
\newcommand{\badgebox}[2]{\begingroup\setlength{\fboxsep}{1pt}\fcolorbox{white}{#1}{\tiny\textsf{#2}}\endgroup}
\newcommand{\scenebadge}{\badgebox{badgeblue}{scene}}
\newcommand{\promptbadge}{\badgebox{badgegreen}{prompt}}
\newcommand{\mixedbadge}{\badgebox{badgeamber}{mixed}}

\begin{table}[t]
\centering
\small
\setlength{\tabcolsep}{3pt}
\renewcommand{\arraystretch}{1.05}
\caption{Intervention taxonomy. Preserving interventions leave task semantics unchanged, whereas semantic-breaking interventions modify action-relevant task semantics. Beige rows provide prompt-edit examples for families that admit an instruction-level realization.}
\label{tab:intervention-taxonomy}
\begin{adjustbox}{max width=\columnwidth}
\begin{tabular}{>{\raggedright\arraybackslash}p{0.23\columnwidth}>{\raggedright\arraybackslash}p{0.29\columnwidth}>{\raggedright\arraybackslash}p{0.44\columnwidth}}
\toprule
\textbf{Type} & \textbf{Family} & \textbf{Representative intervention} \\
\midrule
\rowcolor{lightblueblock}
Preserving & illumination change \scenebadge & darkened scene / lighting shift \\
\rowcolor{lightblueblock}
Preserving & style shift \scenebadge & style-transferred scene variant \\
\rowcolor{lightblueblock}
Preserving & clutter \scenebadge & perimeter clutter around the workspace \\
\rowcolor{lightblueblock}
Preserving & noise \scenebadge & margin noise band / hard noise band \\
\rowcolor{lightblueblock}
Preserving & layout / viewpoint \scenebadge & layout shift / view jitter \\
\rowcolor{lightblueblock}
Preserving & wording control \promptbadge & paraphrase / redundant wording \\
\cdashline{2-3}
\rowcolor{lightbeigeblock}
 & \scriptsize\textit{prompt example} & \makecell[l]{\scriptsize clean: ``put the yellow mug in the microwave''\\[-1pt]\scriptsize edited: ``place the yellow mug into the microwave''} \\
\hdashline
\rowcolor{midblueblock}
Semantic-breaking & target-object swap \mixedbadge & milk $\leftrightarrow$ orange juice \\
\cdashline{2-3}
\rowcolor{lightbeigeblock}
 & \scriptsize\textit{prompt example} & \makecell[l]{\scriptsize clean: ``put the milk in the basket''\\[-1pt]\scriptsize edited: ``put the orange juice in the basket''} \\
\rowcolor{midblueblock}
Semantic-breaking & destination change \mixedbadge & back/front compartment swap \\
\cdashline{2-3}
\rowcolor{lightbeigeblock}
 & \scriptsize\textit{prompt example} & \makecell[l]{\scriptsize clean: ``put the book in the back compartment of the caddy''\\[-1pt]\scriptsize edited: ``put the book in the front compartment of the caddy''} \\
\rowcolor{midblueblock}
Semantic-breaking & spatial-relation change \mixedbadge & left-of / behind placement change \\
\cdashline{2-3}
\rowcolor{lightbeigeblock}
 & \scriptsize\textit{prompt example} & \makecell[l]{\scriptsize clean: ``place the black bowl to the left of the plate''\\[-1pt]\scriptsize edited: ``place the black bowl behind the plate''} \\
\rowcolor{midblueblock}
Semantic-breaking & order change \promptbadge & subgoal order swap \\
\cdashline{2-3}
\rowcolor{lightbeigeblock}
 & \scriptsize\textit{prompt example} & \makecell[l]{\scriptsize clean: ``put the mug in the microwave, then close the door''\\[-1pt]\scriptsize edited: ``close the door, then put the mug in the microwave''} \\
\rowcolor{midblueblock}
Semantic-breaking & executability change \mixedbadge & open/closed state change \\
\cdashline{2-3}
\rowcolor{lightbeigeblock}
 & \scriptsize\textit{prompt example} & \makecell[l]{\scriptsize clean: ``put the bowl in the open drawer''\\[-1pt]\scriptsize edited: ``put the bowl in the closed drawer''} \\
\addlinespace[1pt]
\midrule
\multicolumn{3}{>{\raggedright\arraybackslash}p{\dimexpr0.96\columnwidth\relax}}{\scriptsize\textit{Badge legend.} \scenebadge\ indicates an intervention that primarily changes the scene or observation; \promptbadge\ indicates an instruction-side edit; \mixedbadge\ indicates settings that commonly involve both scene-side and prompt-side changes, while still defining a task-level semantic change rather than a mere rewording.} \\
\bottomrule
\end{tabular}
\end{adjustbox}
\vspace{-0.2cm}
\end{table}

We study reliability of a frozen vision-language-action policy under \emph{controlled task interventions}. The central question is whether \emph{action behavior} changes in the right way when the task description or scene is edited: preserving edits should leave behavior largely unchanged, whereas semantic-breaking edits should induce a clear behavioral split. Let a rollout state at time $t$ be described by an observation $o_t$, a natural-language instruction $x$, and rollout context $h_t$. A frozen base policy $\pi_0 : (o_t, x, h_t) \mapsto a_t^{(0)}$ maps this tuple to an action or action chunk $a_t^{(0)} \in \mathbb{R}^{d_a}$. We write $\mathcal{T}(o,x)$ for the latent task semantics induced by scene and instruction.

A preserving intervention $g \in \mathcal{G}_{\mathrm{pres}}$ produces $(o',x') = g(o,x)$ with $\mathcal{T}(o',x') = \mathcal{T}(o,x)$; examples include illumination, style, clutter, noise, viewpoint, and semantically equivalent rephrasings. A semantic-breaking intervention $b \in \mathcal{G}_{\mathrm{break}}$ produces $(o',x') = b(o,x)$ with $\mathcal{T}(o',x') \neq \mathcal{T}(o,x)$; examples include target-object swaps, destination changes, order changes, and executability changes.

\begin{figure}[t]
\centering
\includegraphics[width=\textwidth]{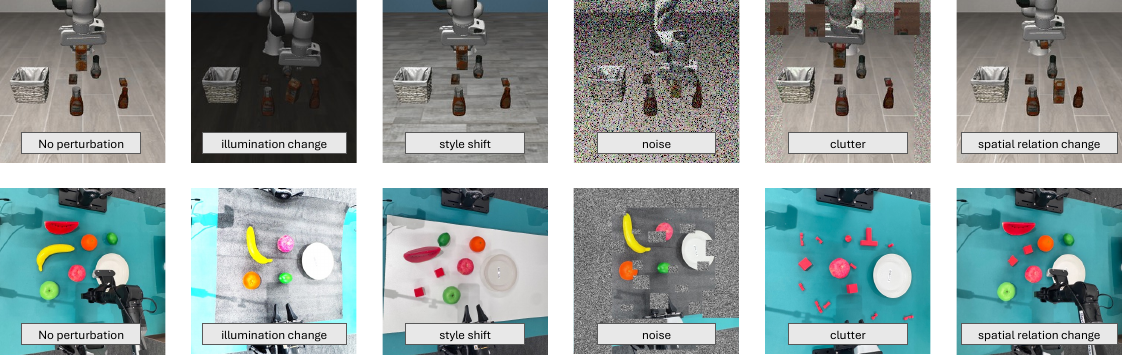}
\caption{Representative intervention families in simulation and real deployment. Top: simulation. Bottom: real deployment. Columns show no perturbation, illumination change, style shift, noise, clutter, and spatial-relation change.}
\label{fig:perturbation-grid}
\end{figure}

Table~\ref{tab:intervention-taxonomy} summarizes the intervention families emphasized here. Some define protocol scope, whereas others receive dedicated quantitative evaluation in the main text or appendix.
Figure~\ref{fig:perturbation-grid} complements this taxonomy with representative simulation and real-deployment views of the same preserving and breaking families.

To distinguish semantic change from harmless variation, we use a three-condition protocol built from a fixed scene and a matched instruction triplet $(x^{c}, x^{u}, x^{b})$, where $x^c$ is the clean instruction, $x^u$ is a semantics-preserving control variant, and $x^b$ is a semantic-breaking variant. For a fixed scene $o$, we require $\mathcal{T}(o,x^c)=\mathcal{T}(o,x^u)$ and $\mathcal{T}(o,x^b)\neq \mathcal{T}(o,x^c)$. We refer to the resulting conditions as \emph{clean}, \emph{control}, and \emph{break}. Throughout the paper, the \emph{Break} column reports clean-task success under the break condition, so lower values indicate stronger stale-task suppression and semantic separation rather than direct changed-task success.

The behavioral requirements are intentionally asymmetric: preserving interventions should leave clean behavior and task success largely intact, whereas semantic-breaking interventions should move the policy away from the clean solution. The primary comparison object is therefore a matched task triplet under a fixed carrier and rollout interface. Task outcome is the main metric, while action-level diagnostics serve only as supporting evidence for whether break conditions leave the clean basin more decisively than preserving controls do. Appendix~\ref{app:protocol-details} summarizes the benchmark-role assignments, rollout budgets, and matched-comparison rules used in the empirical sections.
\vspace{-0.2cm}

\section{Method}
\vspace{-0.2cm}
We instantiate \textsc{BAS-VLA} as a lightweight action-calibration layer on top of a frozen vision-language-action policy. It operates as a lightweight action-calibration layer while leaving the underlying VLA unchanged. It acts at the final action level and asks whether a fixed policy substrate can be selectively recalibrated so that it remains stable under task-preserving changes and separates under genuine task-semantic changes.

Our empirical evidence supports a \emph{selective} design rather than a symmetric two-branch architecture that is always active. The deployed form of BAS-VLA therefore contains two components with different roles: a \emph{breaking-centered calibration core}, which serves as the default method, and a \emph{selective evidence-gated preserving auxiliary}, which is activated only when the rollout remains in an early corrective phase, nuisance evidence is strong, and task semantics remain consistent. The full composition is retained for ablation and mechanism analysis, but it is not treated as the default best-performing variant. Figure~\ref{fig:overview}(B) summarizes this action-calibration view, and Figure~\ref{fig:overview}(C) previews the task-aware visual filtering probe used by the preserving auxiliary.
\vspace{-0.1cm}
\subsection{Overall Formulation}
\vspace{-0.1cm}
Let the frozen base policy be $\pi_0 : (o_t, x, h_t) \mapsto a_t^{(0)}$, where $o_t$ is the observation at time $t$, $x$ is the instruction, $h_t$ is rollout context, and $a_t^{(0)} \in \mathbb{R}^{d_a}$ is the predicted action or action chunk. BAS-VLA outputs a calibrated action $a_t^{\mathrm{BAS}}$ by composing one or two lightweight modules on top of $\pi_0$:
\begin{equation}
a_t^{\mathrm{BAS}}
=
\mathcal{P}_{\phi}\!\left(
\mathcal{B}_{\theta}(a_t^{(0)}, o_t, x, h_t),
\; o_t, x, h_t
\right).
\end{equation}
Here $\mathcal{B}_{\theta}$ denotes the breaking-centered core and $\mathcal{P}_{\phi}$ denotes the preserving auxiliary. In the default deployment, only $\mathcal{B}_{\theta}$ is enabled. The preserving sidecar is not assigned by perturbation family; it activates only when the evidence gate in Eq.~\eqref{eq:pres-gate} is positive. In preserving-only deployment, the auxiliary fuses a probe action with the frozen base action; in the full composition retained for ablation, the same evidence-gated weak-fusion rule is applied on top of the breaking-centered action.
This parameterization reflects the design objective: \emph{selective behavioral calibration}. Preserving interventions should induce small action changes, whereas semantic-breaking interventions should induce large ones.
\vspace{-0.1cm}
\subsection{Breaking-Centered Task-Semantic Calibration}
\vspace{-0.1cm}
The breaking-centered core is the default BAS-VLA component. Its purpose is to preserve clean competence while forcing semantically changed instructions to leave the clean/control action basin. We first compute the base action $a_t^{(0)}=\pi_0(o_t,x,h_t)$ and a frozen instruction feature $z(x)\in\mathbb{R}^{d_z}$. We then define a residual calibrator $r_{\theta}:\mathbb{R}^{d_a}\times\mathbb{R}^{d_z}\rightarrow\mathbb{R}^{d_a}$ and construct
\begin{equation}
a_t^{\mathrm{br}} = a_t^{(0)} + \lambda\, r_{\theta}\!\left(a_t^{(0)}, z(x)\right),
\end{equation}
where $\lambda > 0$ controls the residual scale. This residual form preserves the frozen base policy as the clean-task reference and keeps BAS-VLA portable across carriers because the method modifies action output rather than replacing the whole policy stack.

Training is built from matched instruction triplets $(x^c, x^u, x^b)$, where $x^c$ is the clean instruction, $x^u$ is a semantics-preserving control variant, and $x^b$ is a semantic-breaking variant. The objective enforces three properties simultaneously: clean and control should remain close to the clean reference action, clean and control should remain close to one another, and the breaking action should move farther away from the clean basin than the control action does. Let $a^\star_t$ denote the clean action. We write
\begin{equation}
\begin{aligned}
\mathcal{L}_{\mathrm{imit}}
&=
\|a^{\mathrm{br}}(x^c)-a^\star_t\|_2^2
+
\|a^{\mathrm{br}}(x^u)-a^\star_t\|_2^2,\\
\mathcal{L}_{\mathrm{cons}}
&=
\|a^{\mathrm{br}}(x^c)-a^{\mathrm{br}}(x^u)\|_2^2,\\
\mathcal{L}_{\mathrm{sep}}
&=
\max\!\Big(
0,\;
m
- \|a^{\mathrm{br}}(x^b)-a^{\mathrm{br}}(x^c)\|_2
+ \|a^{\mathrm{br}}(x^u)-a^{\mathrm{br}}(x^c)\|_2
\Big),
\end{aligned}
\end{equation}
where $m>0$ is a separation margin. The final breaking-core objective is
\begin{equation}
\mathcal{L}_{\mathrm{br}}
=
\mathcal{L}_{\mathrm{imit}}
+ \lambda_c \mathcal{L}_{\mathrm{cons}}
+ \lambda_s \mathcal{L}_{\mathrm{sep}}.
\end{equation}
This objective encodes the desired asymmetry directly: preserving rephrasings are contracted toward the clean basin, whereas breaking edits are pushed away from it. During training, the base VLA $\pi_0$ remains frozen and only $r_{\theta}$ is optimized. At inference time, the breaking core adds one frozen instruction feature and one small residual forward pass. Appendix~\ref{app:breaking-properties} states the direct margin consequence of $\mathcal{L}_{\mathrm{sep}}$, and Appendix~\ref{app:residual-bounds} records the corresponding local drift bound induced by the residual form.
\vspace{-0.1cm}
\subsection{Evidence-Gated Preserving Auxiliary}
\begin{figure}[t]
\centering
\includegraphics[width=\textwidth]{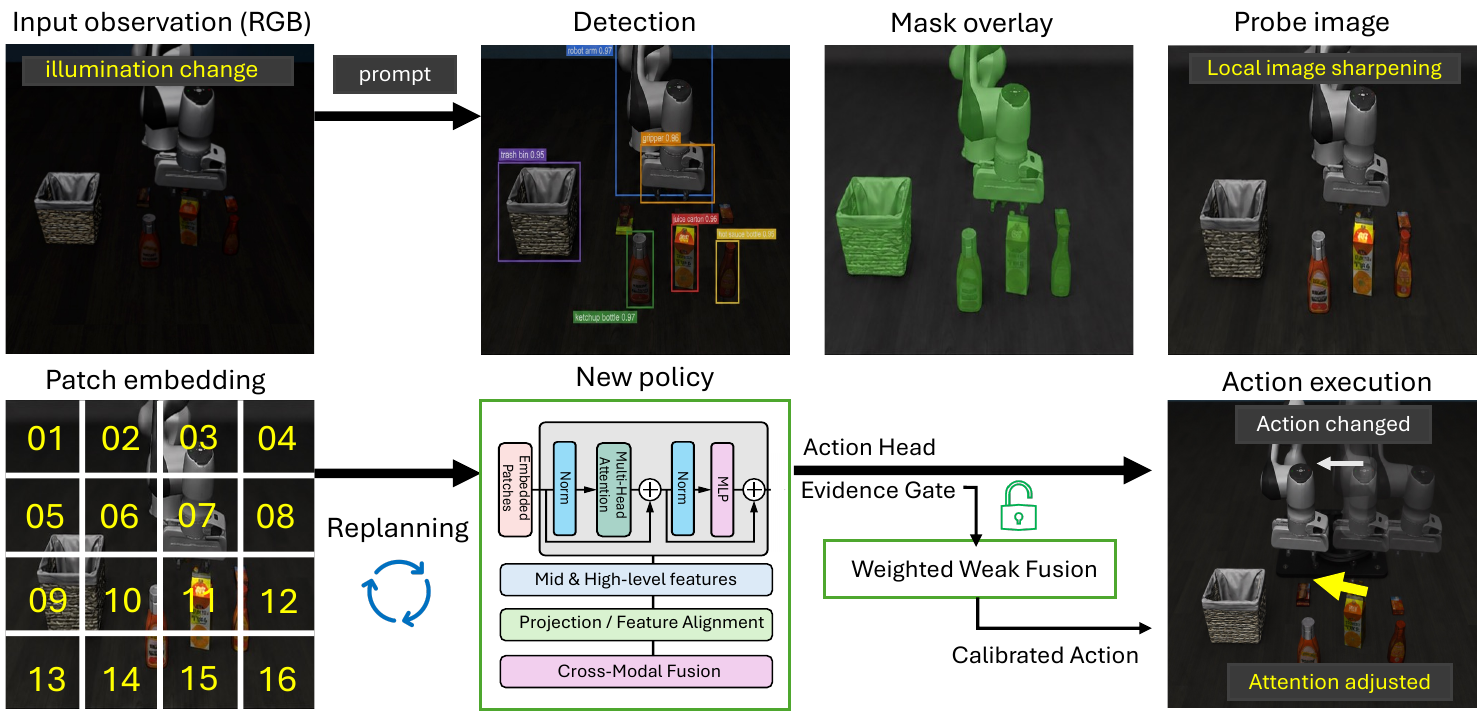}
\caption{Dark-scene preserving example. The frozen carrier misreads the target under illumination change; the evidence-gated probe path activates and produces a corrected action.}
\label{fig:preserving-dark-example}
\end{figure}

\vspace{-0.1cm}
The preserving side of BAS-VLA is deliberately selective, providing a lightweight correction when nuisance-induced action drift is more plausible than a genuine task change. Given the current observation $o_t$, we construct a probe view $\tilde{o}_t^{\mathrm{sty}} = \phi_{\mathrm{sty}}(o_t)$, where the subscript ``sty'' is mnemonic for observation-side visual variation in the input image rather than only the style-shift slice used in one validated benchmark. The transform $\phi_{\mathrm{sty}}$ is a deterministic, non-generative, geometry-preserving appearance-canonicalization map. It suppresses low-level cues such as color cast, illumination bias, and texture-heavy surface variation while preserving object identity, layout, boundaries, and task-relevant geometry. A concrete grounded-mask instantiation of $\phi_{\mathrm{sty}}$, based on text-conditioned grounding and mask refinement~\citep{liu2023groundingdino,ravi2024sam2,he2010guided}, is deferred to Appendix~A.2. We then reuse frozen visual features from the carrier's own visual pathway rather than introducing a separate trainable vision module. Let
\[
v_t^{\mathrm{mid}} = f_{\mathrm{vis}}^{\mathrm{mid}}(o_t), \qquad
\tilde{v}_t^{\mathrm{mid}} = f_{\mathrm{vis}}^{\mathrm{mid}}(\tilde{o}_t^{\mathrm{sty}})
\]
denote a frozen mid-level visual representation, and let
\[
v_t^{\mathrm{late}} = f_{\mathrm{vis}}^{\mathrm{late}}(o_t), \qquad
\tilde{v}_t^{\mathrm{late}} = f_{\mathrm{vis}}^{\mathrm{late}}(\tilde{o}_t^{\mathrm{sty}})
\]
denote a frozen later semantic visual representation from the same backbone. The frozen base policy is then queried on the probe:
\[
a_t^{\mathrm{sty}} = \pi_0(\tilde{o}_t^{\mathrm{sty}}, x, h_t).
\]
Instead of a hard nuisance label, the preserving auxiliary uses an evidence gate
\begin{equation}
\tau_t
=
\tau_{\mathrm{phase}}(t,h_t)\,
g_{\mathrm{vis}}(v_t^{\mathrm{mid}},\tilde{v}_t^{\mathrm{mid}})\,
g_{\mathrm{sem}}(v_t^{\mathrm{late}},\tilde{v}_t^{\mathrm{late}},x)\,
g_{\mathrm{act}}(a_t^{(0)},a_t^{\mathrm{sty}}),
\label{eq:pres-gate}
\end{equation}
where $\tau_{\mathrm{phase}} \in [0,1]$ downweights late-stage intervention, $g_{\mathrm{vis}} \in [0,1]$ measures the strength of the visual shift through frozen mid-level feature discrepancy, $g_{\mathrm{sem}} \in [0,1]$ measures semantic consistency through frozen later visual features, and $g_{\mathrm{act}} \in [0,1]$ measures whether the visual change has already perturbed the base action. Concrete score parameterizations are given in Appendix~B.3. This makes the auxiliary selective in both time and evidence: it should activate only when there is substantial surface variation at the frozen visual-feature level, preserved higher-level semantics, and action-level disagreement that is still early enough to correct.
The auxiliary then applies weak fusion between the base action and the visual-probe action:
\begin{equation}
a_t^{\mathrm{pres}}
=
(1-\alpha\tau_t)a_t^{(0)}+\alpha\tau_t a_t^{\mathrm{sty}},
\end{equation}
with a small fusion weight $\alpha\in(0,1)$. When $\tau_t=0$, this reduces exactly to $a_t^{(0)}$. The choice of weak fusion rather than hard override keeps the frozen base action dominant while allowing a bounded correction when nuisance evidence is present. Figure~\ref{fig:preserving-dark-example} shows one dark-scene instance in which the probe path is activated after a target-reading error under illumination change. Appendix~\ref{app:preserving-properties} makes the resulting gate and drift identities explicit.

\subsection{Deployment Modes}
\vspace{-0.1cm}
This work uses three deployment modes. The default method uses only the breaking-centered core, $a_t^{\mathrm{default}} = a_t^{\mathrm{br}}$, and this is the main variant referred to as BAS-VLA in the semantic-breaking results. For preserving-side stabilization, we deploy the auxiliary through the evidence-gated action
\[
a_t^{\mathrm{pres}}=(1-\alpha\tau_t)a_t^{(0)}+\alpha\tau_t a_t^{\mathrm{sty}}.
\]
For ablation and mechanism analysis, we also retain the full composition
\[
a_t^{\mathrm{full}}=\mathcal{P}_{\phi}(a_t^{\mathrm{br}}, o_t, x, h_t)=(1-\alpha\tau_t)a_t^{\mathrm{br}}+\alpha\tau_t a_t^{\mathrm{sty}}.
\]
Full receives no perturbation-family label and provides a unified composition for mixed conditions. The selective/default modes provide the strongest setting-specific results in our evaluations.
Inference-time cost relations, control-flow details, and scope limits are deferred to Appendix~\ref{app:complexity-scope}, Appendix~\ref{app:deployment-pseudocode}, and Appendix~\ref{app:deployment-properties}.

\begin{table}[t]
\centering
\small
\setlength{\tabcolsep}{2pt}
\caption{Main task-preserving style-shift results on Bowl-on-Ramekin. Each method uses 200 episodes over four seeds; $\Delta_{\mathrm{style}}$ is measured relative to the baseline style-shift row.}
\label{tab:preserving-main}
\begin{tabular}{>{\raggedright\arraybackslash}p{0.13\columnwidth}>{\centering\arraybackslash}p{0.19\columnwidth}>{\centering\arraybackslash}p{0.27\columnwidth}>{\centering\arraybackslash}p{0.21\columnwidth}>{\centering\arraybackslash}p{0.15\columnwidth}}
\toprule
\textbf{Method} & \makecell{\textbf{Clean} \textbf{success} $\uparrow$} & \makecell{\textbf{Matched-clean subset}\\\textbf{success} $\uparrow$} & \makecell{\textbf{Style-shift} \textbf{success} $\uparrow$} & \makecell{$\Delta_{\mathrm{style}}$ $\uparrow$} \\
\midrule
Baseline & \ratecell{194/200}{97.0\%} & \ratecell{108/200}{54.0\%} & \ratecell{84/200}{42.0\%} & --- \\
\rowcolor{lightbeigeblock}
\textbf{BAS-VLA} & \ratecell{195/200}{97.5\%} & \ratecell{108/200}{54.0\%} & \ratecell{140/200}{70.0\%} & $+28.0$ pts \\
\bottomrule
\end{tabular}
\end{table}

\begin{table}[t]
\centering
\footnotesize
\setlength{\tabcolsep}{1pt}
\caption{Main semantic-breaking results on matched target-object triplets. Each row uses 200 clean, 200 control, and 200 break episodes; lower is better in \emph{Break}.}
\label{tab:breaking-main}
\begin{tabular}{>{\raggedright\arraybackslash}p{0.14\columnwidth}>{\raggedright\arraybackslash}p{0.20\columnwidth}>{\centering\arraybackslash}p{0.18\columnwidth}>{\centering\arraybackslash}p{0.18\columnwidth}>{\centering\arraybackslash}p{0.16\columnwidth}>{\centering\arraybackslash}p{0.11\columnwidth}}
\toprule
{\scriptsize\textbf{Task}} & {\scriptsize\textbf{Setting}} & \makecell{\scriptsize\textbf{Clean}\\\scriptsize\textbf{success} $\uparrow$} & \makecell{\scriptsize\textbf{Control}\\\scriptsize\textbf{success} $\uparrow$} & \makecell{\scriptsize\textbf{Break}\\\scriptsize\textbf{(clean)} $\downarrow$} & \makecell{\scriptsize\textbf{Ctrl.-Break}\\\scriptsize\textbf{gap} $\uparrow$} \\
\midrule
\rowcolor{lightgrayblock}
\multicolumn{6}{l}{\textit{Canonical harder target-object tasks}} \\
\rowcolor{lightblueblock}
BBQ-Swap & BBQ sauce$\rightarrow$ketchup & \ratecell{180/200}{90.0\%} & \ratecell{185/200}{92.5\%} & \ratecell{0/200}{0.0\%} & 92.5 \\
\rowcolor{lightblueblock}
Ketchup-Swap & ketchup$\rightarrow$tomato sauce & \ratecell{176/200}{88.0\%} & \ratecell{184/200}{92.0\%} & \ratecell{0/200}{0.0\%} & 92.0 \\
\rowcolor{lightblueblock}
Milk-Swap & milk$\rightarrow$orange juice & \ratecell{196/200}{98.0\%} & \ratecell{195/200}{97.5\%} & \ratecell{0/200}{0.0\%} & 97.5 \\
Butter-Swap & butter$\rightarrow$cream cheese & \ratecell{188/200}{94.0\%} & \ratecell{182/200}{91.0\%} & \ratecell{113/200}{56.5\%} & 34.5 \\
\rowcolor{lightblueblock}
OJ-Swap & orange juice$\rightarrow$milk & \ratecell{197/200}{98.5\%} & \ratecell{193/200}{96.5\%} & \ratecell{0/200}{0.0\%} & 96.5 \\
\addlinespace[1pt]
\hdashline
\addlinespace[1pt]
\rowcolor{lightgrayblock}
\multicolumn{6}{l}{\textit{Stronger interference in the orange-juice$\rightarrow$milk setting}} \\
\rowcolor{midblueblock}
OJ-Swap & Hard noise & \ratecell{184/200}{92.0\%} & \ratecell{185/200}{92.5\%} & \ratecell{0/200}{0.0\%} & 92.5 \\
\rowcolor{deepblueblock}
OJ-Swap & Clutter+Noise & \ratecell{192/200}{96.0\%} & \ratecell{183/200}{91.5\%} & \ratecell{0/200}{0.0\%} & 91.5 \\
\bottomrule
\end{tabular}
\end{table}

\begin{figure}[t]
\centering
\begin{subfigure}[t]{0.3\textwidth}
\centering
\includegraphics[width=\linewidth]{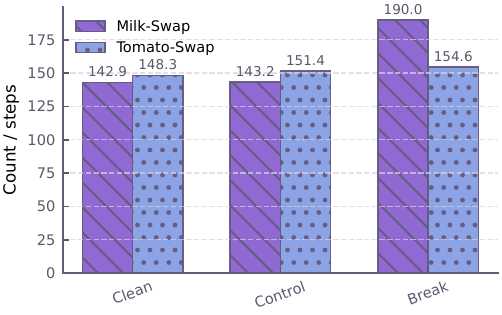}
\caption{Mean steps.}
\end{subfigure}\hfill
\begin{subfigure}[t]{0.3\textwidth}
\centering
\includegraphics[width=\linewidth]{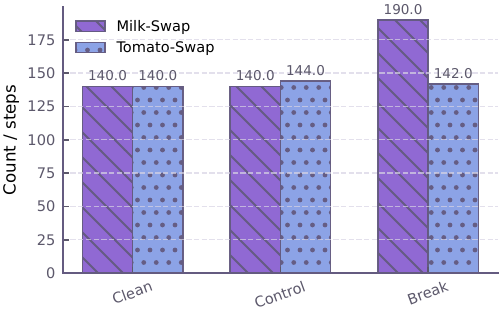}
\caption{Median steps.}
\end{subfigure}\hfill
\begin{subfigure}[t]{0.3\textwidth}
\centering
\includegraphics[width=\linewidth]{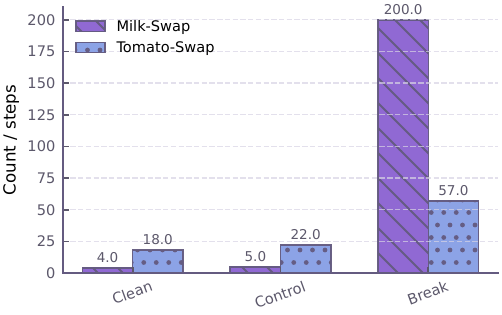}
\caption{190-step saturation.}
\end{subfigure}
\caption{Process-side view of the two main semantic-breaking tasks. Milk-Swap keeps clean/control short and unsaturated while break saturates; Tomato-Swap shows a weaker version of same pattern.}
\label{fig:main-process-panels}
\end{figure}
\vspace{-0.1cm}
\section{Experiments}

We evaluate \textsc{BAS-VLA} along three axes: semantic-breaking reliability, task-preserving stability, and matched-carrier inference-time strategy comparisons. We then use harder tasks and supplementary benchmarks to evaluate transfer.
\vspace{-0.1cm}
\subsection{Experimental Setup}
\vspace{-0.1cm}
\paragraph{Benchmarks, metrics, and scale.}
The primary evaluation uses LIBERO-Object for semantic breaking, LIBERO-Spatial for task preserving, and LIBERO-10 for corroboration~\citep{liu2023libero}; CALVIN, MetaWorld, ManiSkill2, and RoboCasa appear only as supplementary transfer and boundary settings~\citep{mees2022calvin,yu2020metaworld,gu2023maniskill2,nasiriany2024robocasa}. The main evidence comes from two carriers: OpenPI-pi0.5 for semantic-breaking and matched strategy comparisons~\citep{black2025pi05}, and OpenVLA-OFT for task-preserving main results and OFT corroboration~\citep{kim2025oft}. In the latter case, we follow the OFT fine-tuning work of~\citet{kim2025oft} as the carrier reference and use the same recipe as the basis for our own fine-tuning in the simulation and real-robot settings considered here. Preserving evaluations apply semantics-preserving visual or wording perturbations, whereas the main semantic-breaking evaluations use matched clean/control/break triplets whose \emph{Break} column reports clean-task success under the changed task condition; Table~\ref{tab:motivation-gaps} instead reports native pre-method original-versus-perturbed performance.

Task success rate is the primary metric. In preserving settings, the goal is to maintain success under nuisance variation without harming clean-task performance; in semantic-breaking settings, the goal is to keep clean/control success high while adapting to the changed instruction. In the simulation semantic-breaking evaluations, the Break metric reports old-task success under the changed instruction; lower values therefore indicate stronger suppression of stale behavior. Changed-task completion is evaluated separately using mutually exclusive New, Old, and Other/timeout outcomes. Process statistics such as episode length, saturation, and replanning are reported only as supporting evidence. The main preserving campaign evaluates ten LIBERO-Spatial tasks with fifty episodes per task, yielding 500 episodes per variant-condition pair; the main semantic-breaking campaign uses four seeds with fifty episodes per seed, yielding 200 episodes per condition. Appendix~\ref{app:protocol-details} summarizes the benchmark-role assignments, triplet interpretation, rollout budgets, and the real-robot deployment setup. Wilson 95\% intervals for the principal binary outcomes are reported in Table~\ref{tab:uncertainty}.
\vspace{-0.1cm}
%

\subsection{Main Results}
\vspace{-0.1cm}
We begin with the task-preserving style-shift setting. Table~\ref{tab:preserving-main} compares BAS-VLA with the uncalibrated baseline on Bowl-on-Ramekin under style shift. Style-shift success increases from 42.0\% to 70.0\%, while matched-clean performance remains unchanged. Broader preserving-side aggregates and deployment details are deferred to the appendix.

The main semantic-breaking evidence is consolidated in Table~\ref{tab:breaking-main}. The primary Milk-Swap setting shows the intended high-clean/high-control/low-break pattern. OJ-Swap, BBQ-Swap, and Ketchup-Swap reproduce the same high-clean/high-control/low-break pattern, while Butter-Swap and Tomato-Swap remain harder cases. Butter-Swap in particular retains substantial old-task success, an important negative case consistent with confusion between a visually and semantically similar object pair; we report outcome-level error categories in the appendix. Figure~\ref{fig:main-process-panels} shows the same distinction at the rollout level: Milk-Swap stays short and unsaturated on clean/control but saturates under break, indicating strong suppression of the stale clean-task rollout under semantic change, whereas Tomato-Swap shows a weaker version of the same profile.

%

We next compare with alternative inference-time strategies under matched rollout conditions. All methods are instantiated on the same frozen OpenPI carrier and evaluated under the same semantic-breaking setting, so the comparison isolates the strategy layer. Table~\ref{tab:external-retained} shows that BAS-VLA combines the lowest break-column value with high clean/control performance; the alternative methods either leave more residual clean-task success under semantic breaking or reduce clean/control competence more substantially. The corresponding process-side comparison is deferred to Figure~\ref{fig:external-process-panels} in the appendix. We also evaluate BAS-VLA on four real-robot semantic-change cases under the fixed deployment stack summarized in Table~\ref{tab:compute-realrobot-setup}. Table~\ref{tab:realrobot-main} reports success on the \emph{new} instruction, using 80 trials per method and case. BAS-VLA improves changed-task completion across all four cases; Table~\ref{tab:realrobot-original} retains the clean/control and behavioral diagnostics for the two initial tasks.

\begin{table*}[t]
\centering
\small
\setlength{\tabcolsep}{8pt}
\caption{Matched-carrier external strategy comparison on Milk-Swap. Each strategy uses 200 clean, 200 control, and 200 break episodes; lower is better in \emph{Break}.}
\label{tab:external-retained}
\begin{adjustbox}{width=\textwidth}
\begin{tabular}{lcccc}
\toprule
\textbf{Strategy} & \makecell{\textbf{Clean}\\\textbf{success rate} $\uparrow$} & \makecell{\textbf{Control}\\\textbf{success rate} $\uparrow$} & \makecell{\textbf{Break}\\\textbf{(clean)} $\downarrow$} & \makecell{\textbf{Ctrl.-Break}\\\textbf{gap} $\uparrow$} \\
\midrule
\rowcolor{lightbeigeblock}
\textbf{BAS-VLA} & \ratecell{196/200}{98.0\%} & \ratecell{195/200}{97.5\%} & \ratecell{0/200}{0.0\%} & 97.5 \\
\methodvenue{SGAC-AC}{NeurIPS'25}{so2025sgac} & \ratecell{194/200}{97.0\%} & \ratecell{195/200}{97.5\%} & \ratecell{2/200}{1.0\%} & 96.5 \\
\methodvenue{AAC}{CVPR'26}{liang2026aac} & \ratecell{196/200}{98.0\%} & \ratecell{195/200}{97.5\%} & \ratecell{2/200}{1.0\%} & 96.5 \\
\methodvenue{RTC}{NeurIPS'25}{black2025rtc} & \ratecell{197/200}{98.5\%} & \ratecell{192/200}{96.0\%} & \ratecell{4/200}{2.0\%} & 94.0 \\
\methodvenue{PCD}{ICLR'26}{wu2026pcd} & \ratecell{176/200}{88.0\%} & \ratecell{179/200}{89.5\%} & \ratecell{6/200}{3.0\%} & 86.5 \\
\methodvenue{ST4-VLA}{ICLR'26}{st4vla2026} & \ratecell{189/200}{94.5\%} & \ratecell{190/200}{95.0\%} & \ratecell{45/200}{22.5\%} & 72.5 \\
\methodvenue{HAMLET}{ICLR'26}{koo2026hamlet} & \ratecell{88/200}{44.0\%} & \ratecell{86/200}{43.0\%} & \ratecell{0/200}{0.0\%} & 43.0 \\
\bottomrule
\end{tabular}
\end{adjustbox}
\end{table*}

\begin{table}[t]
\centering
\footnotesize
\setlength{\tabcolsep}{3pt}
\caption{Real-robot changed-task completion. Every case uses 80 trials per method (640 changed-condition rollouts in total). T1/T2 include the 17 changed-condition trials per method in Table~\ref{tab:realrobot-original}.}
\label{tab:realrobot-main}
\begin{tabular}{>{\raggedright\arraybackslash}p{\dimexpr0.065\linewidth-2\tabcolsep\relax}>{\raggedright\arraybackslash}p{\dimexpr0.22\linewidth-2\tabcolsep\relax}>{\centering\arraybackslash}p{\dimexpr0.105\linewidth-2\tabcolsep\relax}>{\centering\arraybackslash}p{\dimexpr0.12\linewidth-2\tabcolsep\relax}>{\centering\arraybackslash}p{\dimexpr0.115\linewidth-2\tabcolsep\relax}>{\centering\arraybackslash}p{\dimexpr0.12\linewidth-2\tabcolsep\relax}>{\centering\arraybackslash}p{\dimexpr0.115\linewidth-2\tabcolsep\relax}>{\centering\arraybackslash}p{\dimexpr0.14\linewidth-2\tabcolsep\relax}}
\toprule
\multirow{2}{*}{\scriptsize\textbf{Case}} & \multirow{2}{*}{\scriptsize\textbf{Semantic change}} & \multirow{2}{*}{\scriptsize\textbf{$n$/method}} & \multicolumn{2}{c}{\scriptsize\textbf{Frozen}} & \multicolumn{2}{c}{\scriptsize\textbf{BAS-VLA}} & \multirow{2}{*}{\scriptsize\textbf{$\Delta$ (pts)}} \\
\cmidrule(lr){4-5}\cmidrule(lr){6-7}
& & & {\scriptsize\textbf{Successes}} & {\scriptsize\textbf{SR} $\uparrow$} & {\scriptsize\textbf{Successes}} & {\scriptsize\textbf{SR} $\uparrow$} & \\
\midrule
T1 & Block target & 80 & 19 & {\scriptsize\textcolor{rateblue}{23.8\%}} & 47 & {\scriptsize\textcolor{rateblue}{58.8\%}} & $+35.0$ \\
T2 & Fruit target & 80 & 8 & {\scriptsize\textcolor{rateblue}{10.0\%}} & 31 & {\scriptsize\textcolor{rateblue}{38.8\%}} & $+28.8$ \\
T3 & Receptacle target & 80 & 13 & {\scriptsize\textcolor{rateblue}{16.3\%}} & 52 & {\scriptsize\textcolor{rateblue}{65.0\%}} & $+48.8$ \\
T4 & Spatial relation & 80 & 11 & {\scriptsize\textcolor{rateblue}{13.8\%}} & 36 & {\scriptsize\textcolor{rateblue}{45.0\%}} & $+31.3$ \\
\midrule
\rowcolor{lightblueblock}
\multicolumn{2}{l}{\textit{Aggregate}} & 320 & 51 & {\scriptsize\textcolor{rateblue}{15.9\%}} & 166 & {\scriptsize\textcolor{rateblue}{51.9\%}} & $+35.9$ \\
\bottomrule
\end{tabular}
\end{table}

The same semantic-breaking pattern persists across the additional target-object settings and under stronger nuisance interference, including hard noise and combined clutter-plus-noise conditions. These results indicate that the observed separation is not tied to a single favorable task instance; Appendix~\ref{app:limitations-discussion} and Figure~\ref{fig:main-harder-perimeter} provide the corresponding scope and visual context.
Table~\ref{tab:dual-goal-main} evaluates changed-task completion directly. Across the four easy swaps, BAS-VLA reaches 68.5\% new-task success, while both the frozen carrier and the norm-matched random residual remain below 3\%. This comparison distinguishes instruction-aligned redirection from residual magnitude alone. Butter-Swap remains the harder case, and the mixed visual-interference setting retains changed-task completion alongside stale-task suppression.

\begin{table}[t]
\centering
\footnotesize
\setlength{\tabcolsep}{3pt}
\caption{Changed-task outcome decomposition. New, Old, and Other/timeout are mutually exclusive. Easy swaps aggregate Milk/OJ/BBQ/Ketchup (four tasks $\times$ four seeds $\times$ 50 episodes); each harder setting uses four seeds $\times$ 50 episodes.}
\label{tab:dual-goal-main}
\begin{tabular}{>{\raggedright\arraybackslash}p{\dimexpr0.25\linewidth-2\tabcolsep\relax}>{\raggedright\arraybackslash}p{\dimexpr0.2\linewidth-2\tabcolsep\relax}>{\centering\arraybackslash}p{\dimexpr0.06\linewidth-2\tabcolsep\relax}>{\centering\arraybackslash}p{\dimexpr0.16\linewidth-2\tabcolsep\relax}>{\centering\arraybackslash}p{\dimexpr0.16\linewidth-2\tabcolsep\relax}>{\centering\arraybackslash}p{\dimexpr0.17\linewidth-2\tabcolsep\relax}}
\toprule
{\scriptsize\textbf{Setting}} & {\scriptsize\textbf{Method}} & {\scriptsize\textbf{$n$}} & {\scriptsize\textbf{New} $\uparrow$} & {\scriptsize\textbf{Old} $\downarrow$} & {\scriptsize\textbf{Other/timeout} $\downarrow$} \\
\midrule
\rowcolor{lightgrayblock}
\multicolumn{6}{l}{\textit{Easy target-object swaps: Milk / OJ / BBQ / Ketchup}} \\
Four-task aggregate & Frozen & 800 & \ratecell{18}{2.3\%} & \ratecell{712}{89.0\%} & \ratecell{70}{8.8\%} \\
\rowcolor{lightblueblock}
Four-task aggregate & BAS-VLA & 800 & \ratecell{548}{68.5\%} & \ratecell{0}{0.0\%} & \ratecell{252}{31.5\%} \\
Four-task aggregate & Random residual & 800 & \ratecell{12}{1.5\%} & \ratecell{708}{88.5\%} & \ratecell{80}{10.0\%} \\
\addlinespace[1pt]
\hdashline
\addlinespace[1pt]
\rowcolor{lightgrayblock}
\multicolumn{6}{l}{\textit{Harder target-object and visual-interference settings}} \\
Butter$\rightarrow$cream cheese & BAS-VLA & 200 & \ratecell{53}{26.5\%} & \ratecell{113}{56.5\%} & \ratecell{34}{17.0\%} \\
\rowcolor{midblueblock}
OJ$\rightarrow$milk + vis. interf. & BAS-VLA & 200 & \ratecell{127}{63.5\%} & \ratecell{0}{0.0\%} & \ratecell{73}{36.5\%} \\
\bottomrule
\end{tabular}
\end{table}

External zero-shot evaluation on LIBERO-Plus and LIBERO-PRO tests transfer beyond our constructed interventions, using matched carriers without tuning on either protocol~\citep{fei2025liberoplus,zhou2025liberopro}. Table~\ref{tab:external-benchmarks} shows consistent success-rate gains on both protocols. Non-object semantic changes additionally cover destination, spatial relation, and order (Table~\ref{tab:nonobject-semantic}).

\begin{table}[t]
\centering
\footnotesize
\setlength{\tabcolsep}{3pt}
\caption{External-protocol success without benchmark-specific tuning. LIBERO-Plus uses four categories (light/background/noise/language) $\times$ ten cases $\times$ 120 rollouts per method; LIBERO-PRO uses ten task-perturbation cases $\times$ 120 rollouts per method.}
\label{tab:external-benchmarks}
\begin{tabular}{>{\raggedright\arraybackslash}p{\dimexpr0.19\linewidth-2\tabcolsep\relax}>{\centering\arraybackslash}p{\dimexpr0.1\linewidth-2\tabcolsep\relax}>{\centering\arraybackslash}p{\dimexpr0.13\linewidth-2\tabcolsep\relax}>{\centering\arraybackslash}p{\dimexpr0.13\linewidth-2\tabcolsep\relax}>{\centering\arraybackslash}p{\dimexpr0.11\linewidth-2\tabcolsep\relax}>{\centering\arraybackslash}p{\dimexpr0.13\linewidth-2\tabcolsep\relax}>{\centering\arraybackslash}p{\dimexpr0.11\linewidth-2\tabcolsep\relax}>{\centering\arraybackslash}p{\dimexpr0.1\linewidth-2\tabcolsep\relax}}
\toprule
\multirow{2}{*}{\scriptsize\textbf{Protocol}} & \multirow{2}{*}{\scriptsize\textbf{Cases}} & \multirow{2}{*}{\scriptsize\textbf{$n$/method}} & \multicolumn{2}{c}{\scriptsize\textbf{Baseline}} & \multicolumn{2}{c}{\scriptsize\textbf{BAS-VLA}} & \multirow{2}{*}{\scriptsize\textbf{$\Delta$ (pts)}} \\
\cmidrule(lr){4-5}\cmidrule(lr){6-7}
& & & {\scriptsize\textbf{Successes}} & {\scriptsize\textbf{SR} $\uparrow$} & {\scriptsize\textbf{Successes}} & {\scriptsize\textbf{SR} $\uparrow$} & \\
\midrule
\rowcolor{lightblueblock}
LIBERO-Plus & $4\times10$ & 4,800 & 2,632 & {\scriptsize\textcolor{rateblue}{54.8\%}} & 3,741 & {\scriptsize\textcolor{rateblue}{77.9\%}} & $+23.1$ \\
\rowcolor{lightblueblock}
LIBERO-PRO & 10 & 1,200 & 443 & {\scriptsize\textcolor{rateblue}{36.9\%}} & 857 & {\scriptsize\textcolor{rateblue}{71.4\%}} & $+34.5$ \\
\bottomrule
\end{tabular}
\end{table}

\vspace{-0.2cm}
\subsection{Ablations}
\vspace{-0.1cm}
The main ablation question in this work is whether the individual calibration mechanisms and deployment modes remain distinguishable under protocol-identical comparison. Table~\ref{tab:ablations} therefore focuses on the semantic-breaking setting and compares the frozen carrier baseline, the offline instruction-margin variant, the online semantic-margin variant, the default BAS-VLA core, and the monolithic full composition under the same OpenPI-pi0.5/LIBERO-Object protocol on two target-object settings.
The frozen carrier baseline leaves substantial residual clean-task success under semantic-breaking interventions; the instruction-margin-only and semantic-margin-only variants improve separation; and the default BAS-VLA core is the strongest protocol-identical deployment. The full composition does not dominate the default core, which is consistent with the intended selective role of the preserving auxiliary. The preserving-side deployment comparison and the focused action-space support view are reported in Tables~\ref{tab:ablations-preserving-complete} and~\ref{tab:action-space-support} in the appendix.

Full receives no perturbation-family label. The fixed-policy gate diagnostics, controlled mask-corruption study, and end-to-end latency measurements are reported in Tables~\ref{tab:gate-diagnostics}, \ref{tab:mask-corruption}, and~\ref{tab:sidecar-latency}. These measurements distinguish the default breaking core from the optional preserving sidecar. Clean and semantic-break false activations are 7.5\% and 8.0\%, respectively, while the preserving-shift false-negative rate is 23.5\%. Oracle-routed Full retains 91.5\% task success and is reported separately as an upper-bound diagnostic. Mask-corruption degradation increases from 2.0 points at IoU $\geq0.7$ to 18.5 points with missed-target or random masks. The breaking core adds 47--53 ms at p50; preserving and Full modes have end-to-end p50/p95 latencies of 924/1,048 and 972/1,112 ms, respectively, including the additional frozen-policy query.

Additional evaluations and boundary cases are deferred to Appendix~\ref{app:analysis-boundaries}.

\section{Conclusion}
\vspace{-0.1cm}
This work identifies a reliability gap in vision-language-action policies that generic robustness does not capture. BAS-VLA addresses it with a breaking-centered frozen-policy calibration core and a selective evidence-gated preserving auxiliary. BAS-VLA maintains high clean/control success under semantic-breaking evaluation and improves task-preserving style-shift performance. Stale-task suppression and changed-task completion are distinct outcomes, measured separately in the dual-goal evaluations. Overall, the results support task-semantic action calibration as a complementary mechanism for controlling when VLA behavior should remain stable and when it should change under task-semantic interventions.

\FloatBarrier
\clearpage
\begin{ack}
The authors received no third-party funding or support for this work and declare no competing interests.
\end{ack}
\bibliographystyle{plainnat}
\bibliography{references}

\clearpage
\appendix
\section{Supplementary Results and Visual Evidence}

This appendix provides supplementary quantitative and qualitative results, including perturbation inventories, additional benchmark results, and additional visual evidence from completed experiments.

\paragraph{Licensing.}
BAS-VLA code is released under the MIT license. OpenPI and SAM~2 use Apache~2.0; LIBERO and OpenVLA-OFT use MIT. Other third-party assets retain the licenses and terms specified in their upstream repositories.

\subsection{Compute and Real-Robot Setup}
Table~\ref{tab:compute-realrobot-setup} summarizes the fixed compute and deployment hardware used for the simulation and real-robot experiments. The compute row records the server configuration used for the main experiments, and the remaining rows summarize the shared robot and camera stack used by the real-robot deployment setup.
\begin{table}[H]
\centering
\small
\setlength{\tabcolsep}{6pt}
\caption{Compute and real-robot setup. Depth streams are retained only for recording and diagnosis; model inputs use RGB only.}
\label{tab:compute-realrobot-setup}
\begin{adjustbox}{width=\columnwidth}
\begin{tabular}{ll}
\toprule
\textbf{Component} & \textbf{Configuration} \\
\midrule
\rowcolor{lightblueblock}
Compute server & 8$\times$ NVIDIA RTX 5880 Ada GPUs (48GB VRAM each); single-GPU execution unless noted \\
\rowcolor{lightbeigeblock}
Robot platform & Songling PiPER, single-arm \\
\rowcolor{lightbeigeblock}
Robot control stack & \texttt{piper\_sdk} over CAN \\
\rowcolor{lightbeigeblock}
Cameras & 2$\times$ Orbbec RGB-D input cameras + 1 record-only overview camera \\
\rowcolor{lightbeigeblock}
Model input modality & RGB only \\
\rowcolor{lightbeigeblock}
Camera SDK & OrbbecSDK v1 \\
\bottomrule
\end{tabular}
\end{adjustbox}
\end{table}

\begin{table}[H]
\centering
\footnotesize
\setlength{\tabcolsep}{5pt}
\caption{Real-robot deployment latency diagnostics. Chunk p50 and p95 report per-chunk latency, while Wall-clock p50 reports full-episode duration.}
\label{tab:realrobot-latency}
\begin{adjustbox}{width=\textwidth}
\begin{tabular}{llcccc}
\toprule
\textbf{Task} & \textbf{Method} & \textbf{Chunk p50} & \textbf{Chunk p95} & \textbf{Underrun rate} & \textbf{Wall-clock p50} \\
\midrule
Color-block grouping & baseline & 345 ms & 620 ms & 0.18 & 34.5 s \\
\rowcolor{lightblueblock}
Color-block grouping & \textbf{BAS-VLA default} & 392 ms & 710 ms & 0.27 & 41.8 s \\
Fruit-to-plate & baseline & 368 ms & 655 ms & 0.22 & 43.2 s \\
\rowcolor{lightblueblock}
Fruit-to-plate & \textbf{BAS-VLA default} & 421 ms & 768 ms & 0.31 & 52.6 s \\
\bottomrule
\end{tabular}
\end{adjustbox}
\end{table}

\paragraph{Extended real-robot evaluation and mode costs.}
The four changed-task cases in Table~\ref{tab:realrobot-main} comprise 640 rollouts, with 80 trials per method and case. T1 and T2 include the 17 changed-condition trials per method reported in Table~\ref{tab:realrobot-original}; the latter also gives the clean/control and behavioral diagnostics. T3 and T4 add receptacle-target and spatial-relation changes.

\begin{table}[H]
\centering
\footnotesize
\setlength{\tabcolsep}{4pt}
\caption{Real-robot semantic-adaptation results on two deployment tasks. Each method-task pair uses 16 clean, 17 control, and 17 break episodes. Here \emph{Break} reports success on the changed instruction, and the separation score averages old-target suppression and new-target first-commit.}
\label{tab:realrobot-original}
\begin{adjustbox}{width=\textwidth}
\begin{tabular}{llccccccc}
\toprule
\textbf{Task} &
\textbf{Method} &
\textbf{$n$ (c/ctrl/br)} &
\textbf{Clean} &
\textbf{Control} &
\textbf{Break} &
\makecell{\textbf{Old-target}\\\textbf{suppr.}} &
\makecell{\textbf{New-target}\\\textbf{first-commit}} &
\makecell{\textbf{Sep.}\\\textbf{score}} \\
\midrule
\rowcolor{lightgrayblock}
\multicolumn{9}{l}{\textit{T1: group the specified colored block with its target set}} \\
baseline & Frozen & $16/17/17$ & \ratecell{9/16}{56.3\%} & \ratecell{9/17}{52.9\%} & \ratecell{4/17}{23.5\%} & \ratecell{7/17}{41.2\%} & \ratecell{6/17}{35.3\%} & 38.2 \\
\rowcolor{lightblueblock}
\textbf{BAS-VLA default} & BAS-VLA & $16/17/17$ & \ratecell{10/16}{62.5\%} & \ratecell{10/17}{58.8\%} & \ratecell{7/17}{41.2\%} & \ratecell{12/17}{70.6\%} & \ratecell{11/17}{64.7\%} & 67.6 \\
\addlinespace[2pt]
\rowcolor{lightgrayblock}
\multicolumn{9}{l}{\textit{T2: place the specified fruit onto the plate}} \\
baseline & Frozen & $16/17/17$ & \ratecell{8/16}{50.0\%} & \ratecell{8/17}{47.1\%} & \ratecell{3/17}{17.6\%} & \ratecell{6/17}{35.3\%} & \ratecell{5/17}{29.4\%} & 32.4 \\
\rowcolor{lightblueblock}
\textbf{BAS-VLA default} & BAS-VLA & $16/17/17$ & \ratecell{9/16}{56.3\%} & \ratecell{9/17}{52.9\%} & \ratecell{6/17}{35.3\%} & \ratecell{11/17}{64.7\%} & \ratecell{10/17}{58.8\%} & 61.8 \\
\bottomrule
\end{tabular}
\end{adjustbox}
\end{table}

\begin{table}[tbp]
\centering
\small
\setlength{\tabcolsep}{5pt}
\caption{End-to-end chunk latency by deployment mode. Sidecar and Full totals include grounding, mask refinement, the probe image, and the additional frozen-policy query, measured over at least 200 chunks per task and three timing passes.}
\label{tab:sidecar-latency}
\begin{tabularx}{\linewidth}{@{}Xcc@{}}
\toprule
\textbf{Mode} & \textbf{T1 p50/p95 (ms)} & \textbf{T2 p50/p95 (ms)} \\
\midrule
Frozen carrier & 345/620 & 368/655 \\
\rowcolor{lightblueblock}
Default breaking core & 392/710 & 421/768 \\
Preserving sidecar & 924/1,048 & 924/1,048 \\
Full composition & 972/1,112 & 972/1,112 \\
\bottomrule
\end{tabularx}
\end{table}

The breaking core adds 47--53 ms at p50. Preserving and Full modes compute the probe action before the action-disagreement factor $g_{\mathrm{act}}$, so the gate controls the probe's influence on execution while the additional policy query remains included in the measured latency.

\FloatBarrier

\subsection{Protocol and Extended Results}
\subsubsection{Protocol Details}
\label{app:protocol-details}

The main semantic-breaking evaluations use matched clean/control/break triplets. Clean denotes the original task condition, control denotes a semantics-preserving variant of that task, and break denotes a task-semantic change. Unless explicitly stated otherwise, the \emph{Break} column in the simulation semantic-breaking tables reports clean-task success under the changed task condition, so lower values indicate stronger semantic separation. Task outcome is the primary metric, and action-level diagnostics are reported only as supporting evidence for whether break conditions move farther from the clean action basin than preserving controls do.

The benchmark and carrier roles are likewise fixed. OpenPI-pi0.5 with LIBERO-Object is the main semantic-breaking and matched external-strategy setting; OpenVLA-OFT with LIBERO-Spatial is the main task-preserving setting; and LIBERO-10 provides supplementary corroboration. CALVIN, MetaWorld, ManiSkill2, and RoboCasa appear only as boundary or transfer settings. The primary preserving campaign evaluates ten LIBERO-Spatial tasks with fifty episodes per task, yielding 500 episodes per variant-condition pair. The primary semantic-breaking campaign evaluates four seeds with fifty episodes per seed, yielding 200 episodes per condition. External strategies are compared only when carrier, benchmark slice, rollout interface, and evaluation budget are matched.

These comparisons are intentionally anchored to strong in-domain carriers rather than weak generic zero-shot policies. OpenPI-pi0.5 is already adapted to LIBERO object manipulation, and OpenVLA-OFT is already adapted to LIBERO-Spatial. The matched-carrier protocol therefore isolates action calibration on competent frozen substrates rather than conflating it with gross backbone mismatch.

The same appendix also records details that are compressed in the main paper. The primary metric is task success rate. In preserving settings, the goal is to maintain success under nuisance variation without harming clean-task performance; in semantic-breaking settings, the goal is to keep clean/control success high while reducing clean-task success under the changed task condition. Process statistics such as episode length, saturation, replanning, and action-gap diagnostics are reported only as supporting evidence. The real-robot deployment study uses the same BAS-VLA interface and is summarized by Tables~\ref{tab:compute-realrobot-setup} and~\ref{tab:realrobot-latency}.

\paragraph{Changed-task and non-object evaluation.}
The dual-goal evaluator assigns each changed-condition episode to mutually exclusive New, Old, and Other/timeout outcomes. Table~\ref{tab:dual-goal-main} reports these outcomes together with the norm-matched random-residual control. Table~\ref{tab:nonobject-semantic} extends evaluation beyond object substitutions; each family uses two tasks, four seeds, and fifty episodes per seed.

\begin{table}[tbp]
\centering
\small
\setlength{\tabcolsep}{7pt}
\caption{Changed-task completion on non-object semantic interventions.}
\label{tab:nonobject-semantic}
\begin{tabular}{lccc}
\toprule
\textbf{Semantic change} & \textbf{Evaluation volume} & \textbf{Metric} & \textbf{BAS-VLA result} \\
\midrule
\rowcolor{lightblueblock}
Destination & $2\times4\times50=400$ & New-task SR & \ratecell{224/400}{56.0\%} \\
\rowcolor{midblueblock}
Spatial relation & $2\times4\times50=400$ & New-task SR & \ratecell{168/400}{42.0\%} \\
\rowcolor{deepblueblock}
Subgoal order & $2\times4\times50=400$ & New-task SR & \ratecell{128/400}{32.0\%} \\
\bottomrule
\end{tabular}
\end{table}

\paragraph{Unified gate and probe robustness.}
A fixed Full policy is evaluated in a $2\times2$ design crossing preserved/changed semantics with absent/present visual shift. The gate target is whether the probe action is closer than the base action to the expert/reference action. Standard Full receives no oracle family label. Table~\ref{tab:gate-diagnostics} reports condition-level false activation and false negatives, with oracle-routed task retention reported separately as an upper-bound diagnostic.

\begin{table}[tbp]
\centering
\small
\setlength{\tabcolsep}{7pt}
\caption{Full-mode gate diagnostics. Every row uses four seeds and 50 episodes per seed.}
\label{tab:gate-diagnostics}
\begin{tabular}{lcc}
\toprule
\textbf{Gate diagnostic} & \textbf{Evaluation volume} & \textbf{Result} \\
\midrule
Clean false activation & 200 & \ratecell{15/200}{7.5\%} \\
Preserving-shift false negative & 200 & \ratecell{47/200}{23.5\%} \\
Semantic-break false activation & 200 & \ratecell{16/200}{8.0\%} \\
\rowcolor{lightgrayblock}
Oracle-routed Full task retention (upper bound) & 200 & \ratecell{183/200}{91.5\%} \\
\bottomrule
\end{tabular}
\end{table}

Controlled mask corruption measures how localization quality propagates through evidence-gated fusion. Table~\ref{tab:mask-corruption} reports preserving-SR changes relative to normal masks. Moderate localization errors cause smaller degradation than missed-target or random masks. The identity $\tau=0\Rightarrow a^{\mathrm{pres}}=a^{(0)}$ describes fallback at a low gate, while these measurements quantify error propagation when the probe influences the action.

\begin{table}[tbp]
\centering
\small
\setlength{\tabcolsep}{7pt}
\caption{Preserving-SR change relative to the normal-mask reference. Each corruption level uses four seeds and 50 episodes per seed.}
\label{tab:mask-corruption}
\begin{tabular}{lcc}
\toprule
\textbf{Mask condition} & \textbf{Evaluation volume} & \textbf{Preserving-SR change} \\
\midrule
\rowcolor{lightblueblock}
IoU $\geq 0.7$ & 200 & $-4/200$ ($-2.0$ pts) \\
\rowcolor{midblueblock}
IoU $=0.5$ & 200 & $-19/200$ ($-9.5$ pts) \\
\rowcolor{deepblueblock}
Missed target or random mask & 200 & $-37/200$ ($-18.5$ pts) \\
\bottomrule
\end{tabular}
\end{table}

\paragraph{Statistical uncertainty.}
Table~\ref{tab:uncertainty} reports Wilson 95\% confidence intervals computed from aggregate binary episode counts. The intervals quantify uncertainty in both the preserving improvement and the high-clean/high-control/low-old-task pattern without requiring episode pairing.

\begin{table}[tbp]
\centering
\footnotesize
\setlength{\tabcolsep}{6pt}
\caption{Wilson 95\% confidence intervals for the principal binary outcomes.}
\label{tab:uncertainty}
\begin{tabular}{lcc}
\toprule
\textbf{Result} & \textbf{Success rate} & \textbf{Wilson 95\% CI} \\
\midrule
Baseline style & \ratecell{84/200}{42.0\%} & $[35.4,48.9]$ \\
\rowcolor{lightbeigeblock}
BAS-VLA style & \ratecell{140/200}{70.0\%} & $[63.3,75.9]$ \\
Milk clean & \ratecell{196/200}{98.0\%} & $[95.0,99.2]$ \\
Milk control & \ratecell{195/200}{97.5\%} & $[94.3,98.9]$ \\
Milk old-task under break & \ratecell{0/200}{0.0\%} & $[0.0,1.9]$ \\
Butter old-task under break & \ratecell{113/200}{56.5\%} & $[49.6,63.2]$ \\
\bottomrule
\end{tabular}

\end{table}

\subsubsection{Complexity, Scope, and Boundaries}
\label{app:complexity-scope}

The BAS-VLA breaking core remains lightweight relative to the underlying policy, adding one instruction feature and one residual module. Preserving and Full modes require an additional base-policy query to compute the probe action before evaluating the action-disagreement gate; the gate controls the influence of that action. Appendix~\ref{app:deployment-pseudocode} summarizes the inference-time control flow, and Appendix~\ref{app:deployment-properties} records the corresponding per-mode cost relations and exact identity relations.

The scope claim is deliberately narrow. BAS-VLA is not presented as a universal preserving branch or a general-purpose robustness layer. The strongest evidence comes from breaking-centered calibration, with selective preserving gains on validated nuisance families. Additional evaluations on other carriers, suites, and local protocols are summarized in Table~\ref{tab:support-corroboration}. Boundary and negative cases, including executability changes and supplementary benchmarks such as MetaWorld, ManiSkill2, RoboCasa, and CALVIN, are summarized in Table~\ref{tab:boundary-negative} and the extended appendix inventories.

\subsubsection{Analysis and Boundary Summaries}
\label{app:analysis-boundaries}
\begin{table}[!htbp]
\centering
\footnotesize
\setlength{\tabcolsep}{2pt}
\caption{Task-preserving perturbation inventory. Baseline-only rows report changed-condition success without BAS-VLA calibration; calibrated rows report the corresponding method-level changed-condition success.}
\label{tab:preserving-inventory}
\begin{tabular}{>{\raggedright\arraybackslash}p{0.08\columnwidth}>{\raggedright\arraybackslash}p{0.13\columnwidth}>{\raggedright\arraybackslash}p{0.11\columnwidth}>{\raggedright\arraybackslash}p{0.12\columnwidth}>{\raggedright\arraybackslash}p{0.11\columnwidth}>{\centering\arraybackslash}p{0.14\columnwidth}>{\centering\arraybackslash}p{0.14\columnwidth}>{\centering\arraybackslash}p{0.09\columnwidth}}
\toprule
{\scriptsize\textbf{Carrier}} & {\scriptsize\textbf{Suite}} & {\scriptsize\textbf{Task slice}} & {\scriptsize\textbf{Perturbation}} & {\scriptsize\textbf{Best method}} & {\scriptsize\textbf{Baseline} $\uparrow$} & {\scriptsize\textbf{Calibrated} $\uparrow$} & \makecell{\scriptsize$\Delta \uparrow$ \scriptsize(pts)} \\
\midrule
\rowcolor{lightgrayblock}
\multicolumn{8}{l}{\textit{Limited-recovery appearance family}} \\
Pi0.5 & LIBERO-Object & OJ-Swap$\times$200 & illumination change & baseline-only & \ratecell{192/200}{96.0\%} & --- & --- \\
\addlinespace[2pt]
\rowcolor{lightgrayblock}
\multicolumn{8}{l}{\textit{Moderate support under stronger nuisance}} \\
\rowcolor{lightblueblock}
OFT & LIBERO-Spatial & all10$\times$20 & clutter+noise & PARR & \ratecell{165/200}{82.5\%} & \ratecell{174/200}{87.0\%} & $+4.5$ \\
\rowcolor{lightblueblock}
OFT & LIBERO-Spatial & all10$\times$20 & hard noise band & PARR & \ratecell{171/200}{85.5\%} & \ratecell{180/200}{90.0\%} & $+4.5$ \\
\addlinespace[2pt]
\rowcolor{lightgrayblock}
\multicolumn{8}{l}{\textit{Style-sensitive families with strongest BAS-VLA gains}} \\
\rowcolor{lightblueblock}
OFT & LIBERO-Spatial & all10$\times$50 & style shift & BAS-VLA (style-aux) & \ratecell{470/500}{94.0\%} & \ratecell{482/500}{96.4\%} & $+2.4$ \\
\rowcolor{midblueblock}
OFT & LIBERO-Spatial & Bowl-on-Ramekin\allowbreak$\times$200 & style shift & BAS-VLA (style-aux) & \ratecell{84/200}{42.0\%} & \ratecell{140/200}{70.0\%} & $+28.0$ \\
\rowcolor{midblueblock}
OFT & LIBERO-Spatial & Bowl-next-to-Plate$\times$200 & style shift & BAS-VLA (style-aux) & \ratecell{96/200}{48.0\%} & \ratecell{138/200}{69.0\%} & $+21.0$ \\
\rowcolor{midblueblock}
OFT & LIBERO-Spatial & Bowl-on-Cabinet$\times$200 & style shift & BAS-VLA (style-aux) & \ratecell{101/200}{50.5\%} & \ratecell{147/200}{73.5\%} & $+23.0$ \\
\rowcolor{deepblueblock}
OFT & LIBERO-Spatial & aggregation & style shift & PSSC & \ratecell{190/200}{95.0\%} & \ratecell{200/200}{100.0\%} & $+5.0$ \\
\bottomrule
\end{tabular}
\end{table}
The main text focuses on the preserving style-shift setting, the primary semantic-breaking rows, and the matched strategy comparison. The appendix collects the broader scope checks: additional evaluations on other carriers and suites, supplementary harder-task or interference results, boundary cases, and negative cases. These rows are included to delimit where the current deployment is supported, where it remains only partially supported, and which benchmark transfers should still be interpreted cautiously.

The headline semantic-breaking rows should not be read as a generic robustness objective. A uniformly robust policy would preserve clean-task success across clean, control, and break alike. Our protocol instead treats low clean-task success under a genuine break as evidence of the intended behavioral split. The process-side panels and matched strategy comparisons are included for the same reason: they show that semantic separation is visible not only in final task outcomes, but also in rollout duration and strategy-level budget use.

\subsubsection{Extended Related Positioning}
\label{app:extended-related}

The related-work positioning in the main text emphasizes only the closest conceptual neighborhoods. The additional policy-side context is that recent inference-time VLA methods differ primarily in what they optimize at test time: RTC and adaptive chunking methods target execution freshness and latency, ST4VLA strengthens spatial grounding, PCD contrasts original and masked observations, and HAMLET or MG-Select emphasize history-aware prediction or selection. These methods are close to BAS-VLA in deployment regime, but not in comparison object: BAS-VLA is organized around controlled preserving and semantic-breaking interventions rather than around generic test-time improvement.

The additional reasoning-side context is that counterfactual feedback, semantic-intervention benchmarks, and process supervision collectively motivate intervention-sensitive evaluation rather than fixed-answer evaluation alone. BAS-VLA imports that logic into embodied control by asking whether action behavior remains invariant when meaning is unchanged and separates when meaning truly changes.

Table~\ref{tab:preserving-inventory} reports task-preserving results across perturbation families and task settings. The inventory is organized to distinguish limited-recovery appearance families, moderate support under stronger nuisance settings, and the strongest style-sensitive BAS-VLA gains. The OpenPI-pi0.5 illumination row serves as an appearance-family reference, while the style-shift rows emphasize the task-specific slices on which the preserving auxiliary is most effective.

\begin{table*}[tbp]
\centering
\small
\setlength{\tabcolsep}{5pt}
\caption{Protocol-identical preserving-side and deployment ablations on the Bowl-on-Ramekin style-shift setting, with the corresponding OFT-LIBERO replay row. Each Bowl-on-Ramekin row uses 200 episodes.}
\label{tab:ablations-preserving-complete}
\begin{adjustbox}{width=\textwidth}
\begin{tabular}{llcccc}
\toprule
\textbf{Variant} &
\textbf{Role} &
\makecell{\textbf{Bowl-on-Ramekin}\\\textbf{clean} $\uparrow$} &
\makecell{\textbf{Bowl-on-Ramekin}\\\textbf{style-shift} $\uparrow$} &
\makecell{$\Delta_{\mathrm{style}}$\\\textbf{vs. baseline} $\uparrow$} &
\makecell{\textbf{OFT-LIBERO}\\\textbf{replay}} \\
\midrule
Carrier baseline & frozen-policy ref.
& \ratecell{108/200}{54.0\%} & \ratecell{84/200}{42.0\%} & --- & native $20/20$ \\

Semantic-margin only & break-only cal.
& \ratecell{108/200}{54.0\%} & \ratecell{83/200}{41.5\%} & $-0.5$ pts & $20/20$ \\

Style-aux only & preserving-only aux.
& \ratecell{107/200}{53.5\%} & \ratecell{132/200}{66.0\%} & $+24.0$ pts & $19/20$ \\

\rowcolor{midblueblock}
\textbf{BAS-VLA (style-aux)} & selective preserve.
& \ratecell{108/200}{54.0\%} & \ratecell{140/200}{70.0\%} & $+28.0$ pts & $20/20$ \\

\rowcolor{deepblueblock}
\textbf{BAS-VLA (full)} & full composition
& \ratecell{106/200}{53.0\%} & \ratecell{134/200}{67.0\%} & $+25.0$ pts & $20/20$ \\
\bottomrule
\end{tabular}
\end{adjustbox}
\end{table*}

\begin{table*}[tbp]
\centering
\small
\setlength{\tabcolsep}{4pt}
\definecolor{rolebadgered}{RGB}{225,120,120}
\definecolor{rolebadgegreen}{RGB}{117,177,126}
\definecolor{rolebadgepurple}{RGB}{156,126,204}
\newcommand{\roleRef}{\raisebox{-0.30ex}{\textcolor{rolebadgered}{\Large$\bullet$}}}
\newcommand{\roleMech}{\raisebox{-0.30ex}{\textcolor{rolebadgegreen}{\Large$\bullet$}}}
\newcommand{\roleDeploy}{\raisebox{-0.30ex}{\textcolor{rolebadgepurple}{\Large$\bullet$}}}
\caption{Protocol-identical breaking-side ablations on two target-object settings. Each clean/control/break column uses 200 episodes; lower is better in \emph{Break}.}
\label{tab:ablations}
\begin{adjustbox}{width=\textwidth}
\begin{tabular}{llcccccccc}
\toprule
\textbf{Variant} &
\textbf{Role} &
\multicolumn{4}{c}{\textbf{Milk-Swap setting}} &
\multicolumn{4}{c}{\textbf{OJ-Swap setting}} \\
\cmidrule(lr){3-6}\cmidrule(lr){7-10}
&
&
\makecell{\textbf{Clean}\\\textbf{$\uparrow$}} &
\makecell{\textbf{Control}\\\textbf{$\uparrow$}} &
\makecell{\textbf{Break}\\\textbf{$\downarrow$}} &
\makecell{\textbf{Gap}\\\textbf{$\uparrow$}} &
\makecell{\textbf{Clean}\\\textbf{$\uparrow$}} &
\makecell{\textbf{Control}\\\textbf{$\uparrow$}} &
\makecell{\textbf{Break}\\\textbf{$\downarrow$}} &
\makecell{\textbf{Gap}\\\textbf{$\uparrow$}} \\
\midrule
Carrier baseline & \roleRef
& \ratecell{196/200}{98.0\%} & \ratecell{194/200}{97.0\%} & \ratecell{180/200}{90.0\%} & 7.0
& \ratecell{197/200}{98.5\%} & \ratecell{193/200}{96.5\%} & \ratecell{176/200}{88.0\%} & 8.5 \\

Instruction-margin only & \roleMech
& \ratecell{196/200}{98.0\%} & \ratecell{195/200}{97.5\%} & \ratecell{72/200}{36.0\%} & 61.5
& \ratecell{197/200}{98.5\%} & \ratecell{192/200}{96.0\%} & \ratecell{80/200}{40.0\%} & 56.0 \\

Semantic-margin only & \roleMech
& \ratecell{196/200}{98.0\%} & \ratecell{195/200}{97.5\%} & \ratecell{16/200}{8.0\%} & 89.5
& \ratecell{196/200}{98.0\%} & \ratecell{192/200}{96.0\%} & \ratecell{22/200}{11.0\%} & 85.0 \\

\rowcolor{lightblueblock}
\textbf{BAS-VLA (core)} & \roleDeploy
& \ratecell{196/200}{98.0\%} & \ratecell{195/200}{97.5\%} & \ratecell{0/200}{0.0\%} & 97.5
& \ratecell{197/200}{98.5\%} & \ratecell{193/200}{96.5\%} & \ratecell{0/200}{0.0\%} & 96.5 \\

\rowcolor{deepblueblock}
\textbf{BAS-VLA (full)} & \roleDeploy
& \ratecell{195/200}{97.5\%} & \ratecell{193/200}{96.5\%} & \ratecell{8/200}{4.0\%} & 92.5
& \ratecell{196/200}{98.0\%} & \ratecell{191/200}{95.5\%} & \ratecell{10/200}{5.0\%} & 90.5 \\
\addlinespace[2pt]
\multicolumn{10}{c}{\scriptsize \roleRef\ frozen-policy reference \hspace{1.2em} \roleMech\ mechanism-only ablation \hspace{1.2em} \roleDeploy\ BAS-VLA deployment variant} \\
\bottomrule
\end{tabular}
\end{adjustbox}
\end{table*}

\subsection{Supplementary Visual Evidence}

This subsection groups the screenshot-based supplementary evidence into three roles. The first role is preserving-side motivation and nuisance robustness, where clean rollouts are juxtaposed with semantics-preserving perturbations to show when the visual change should leave the task unchanged but still induces behavioral drift or failure. The second role is semantic-breaking and harder-setting evidence, where matched clean and break rollouts are extended to stronger interference conditions so that the clean/control/break distinction remains visible at the rollout level. The third role is external-strategy qualitative evidence, whose purpose is not to replace the matched-carrier quantitative tables but to make the behavioral differences between comparison strategies easier to inspect.

\begin{figure}[!htbp]
\centering
\includegraphics[width=\textwidth]{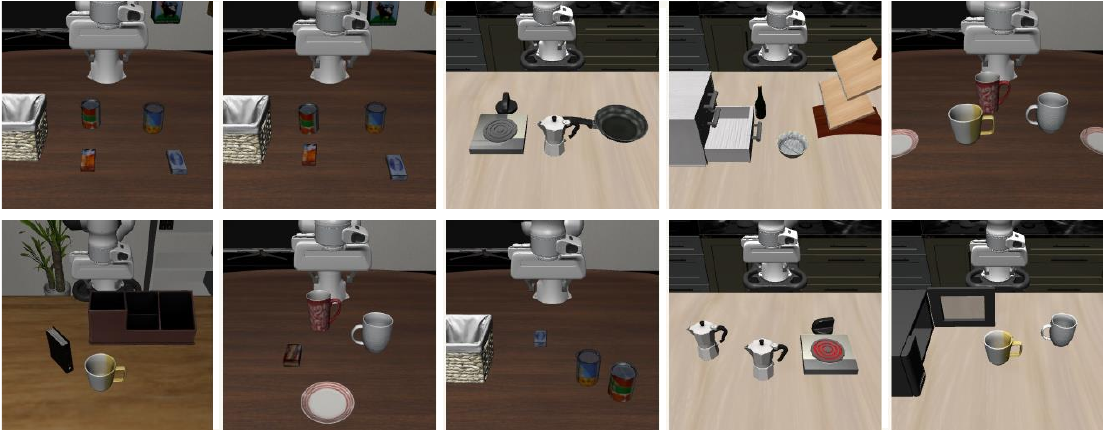}
\caption{Main simulation task scenes. The grid shows the ten task layouts used in the primary simulation campaigns.}
\label{fig:appendix-task-scenes}
\end{figure}

\begin{figure}[!htbp]
\centering
\includegraphics[width=\textwidth]{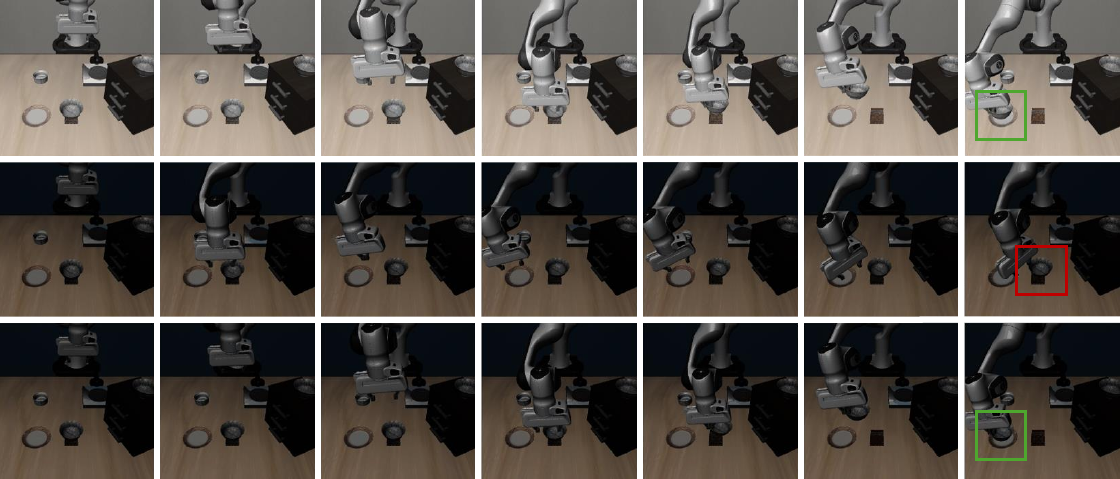}
\caption{Dark-scene preserving comparison. Top: clean success. Middle: illumination-change failure under the frozen baseline. Bottom: the same darkened condition succeeds with BAS-VLA.}
\label{fig:appendix-dark-compare}
\end{figure}

\begin{figure}[!htbp]
\centering
\includegraphics[width=0.99\textwidth]{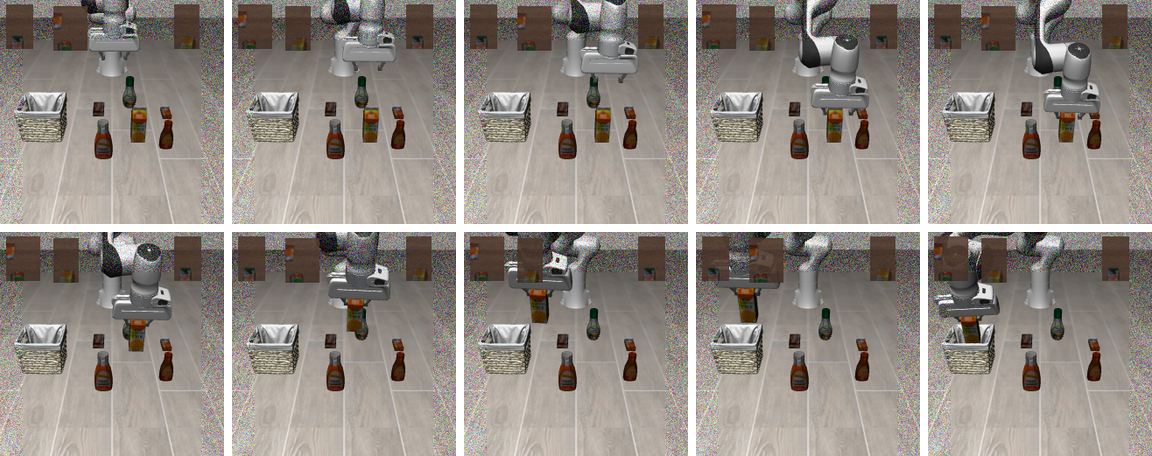}
\caption{Representative OJ-Swap under clutter+noise; BAS-VLA completes the intended task.}
\label{fig:main-harder-perimeter}
\end{figure}

Butter$\rightarrow$cream cheese yields 53/200 New, 113/200 Old, and 34/200 Other/timeout outcomes. Stale-task completion is the dominant failure outcome. The rollout annotations distinguish stale-target grasp, correct-target reach followed by grasp failure, third-object commitment, oscillation, and timeout, separating target confusion from downstream control failure.

\begin{figure}[!htbp]
\centering
\includegraphics[width=\textwidth]{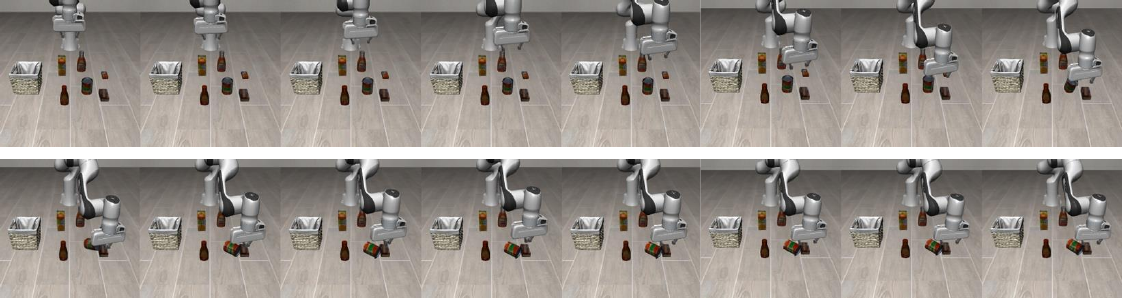}
\caption{Butter-Swap failure sequence. The rollout continues toward the stale clean-task object rather than committing to the swapped target.}
\label{fig:appendix-butter-swap-fail}
\end{figure}

\begin{figure}[!htbp]
\centering
\begin{subfigure}[t]{0.27\textwidth}
\centering
\includegraphics[width=\linewidth]{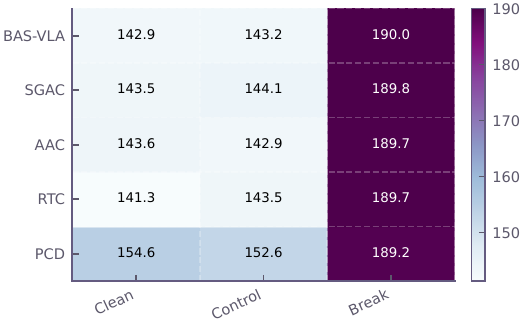}
\caption{Mean steps $\downarrow$.}
\end{subfigure}\hfill
\begin{subfigure}[t]{0.27\textwidth}
\centering
\includegraphics[width=\linewidth]{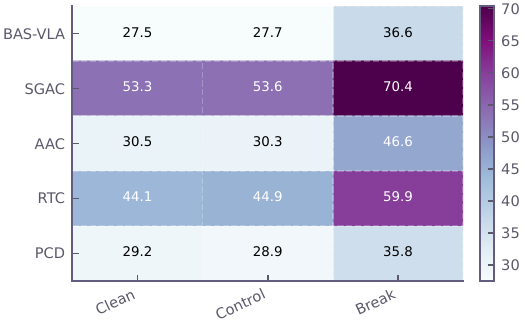}
\caption{Mean replans $\downarrow$.}
\end{subfigure}\hfill
\begin{subfigure}[t]{0.27\textwidth}
\centering
\includegraphics[width=\linewidth]{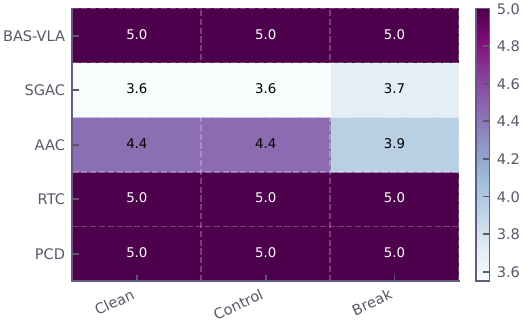}
\caption{Mean selected replan steps $\uparrow$.}
\end{subfigure}
\caption{Process-side comparison among external strategies with a BAS-VLA reference row. Rows are strategies and columns are clean, control, and break.}
\label{fig:external-process-panels}
\end{figure}


\subsection{Concrete Probe Instantiation}

The main text defines the preserving-side probe transform $\phi_{\mathrm{sty}}$ functionally, as a deterministic, non-generative, geometry-preserving appearance-canonicalization map. Here the subscript ``sty'' denotes generic observation-side visual variation in the input image rather than only a narrow style-transfer setting. A concrete instantiation consistent with that definition is a grounded-foreground probe. Given the current observation $o_t$ and instruction $x$, we first use text-conditioned grounding to localize the robot arm, gripper, target object, and task-relevant receptacle cues implied by $x$. We then refine these grounded regions into a merged foreground mask and attenuate only the complement region. The resulting probe leaves task-relevant foreground geometry intact while weakening nuisance-heavy background evidence.

One explicit form is
\[
B_t=\mathrm{GroundDINO}(o_t,x),\qquad
M_t=\mathrm{SAM2}(o_t,B_t),\qquad
\tilde{o}_t^{\mathrm{sty}}=M_t\odot o_t + (1-M_t)\odot \psi_{\mathrm{bg}}(o_t),
\]
where $B_t$ denotes grounded boxes, $M_t$ denotes the merged foreground mask, and $\psi_{\mathrm{bg}}$ denotes deterministic background attenuation on the mask complement. In this work, $\psi_{\mathrm{bg}}$ is intended to reduce color cast, illumination bias, and texture-heavy surface variation without altering object layout or semantic structure.

\subsection{Support and Boundary Analyses}
This subsection complements the main-text results with additional evaluations that clarify scope. We separate them into two groups. The first group consists of \emph{support and corroboration} evaluations from additional suites and carriers. These evaluations report whether the same calibration logic remains visible once the benchmark family, carrier, or local success criterion changes. The second group consists of \emph{boundary and negative} evaluations covering benchmark-specific, protocol-specific, or integration-heavy cases.

Table~\ref{tab:support-corroboration} reports additional results beyond the main target-object protocol. The table is organized into direct triplet corroboration on additional semantic-breaking families, matched support comparisons on preserving-side settings, and native corroboration rows reported under local benchmark metrics. These rows are supportive rather than protocol-identical to the main comparison setting, and are included to show whether the qualitative direction of the BAS-VLA effect remains visible once the carrier, suite, or local metric changes.
\begin{table}[tbp]
\centering
\small
\setlength{\tabcolsep}{7pt}
\caption{Support and corroboration results across additional suites and carriers. Triplet rows report clean, control, and break directly; matched-support and native-corroboration rows state their local comparison explicitly.}
\label{tab:support-corroboration}
\begin{adjustbox}{width=\textwidth}
\begin{tabular}{lllll}
\toprule
\textbf{Slice} & \textbf{Carrier} & \textbf{Suite} & \textbf{Protocol} & \textbf{Result} \\
\midrule
\rowcolor{lightblueblock}
\multicolumn{5}{l}{\textit{Direct triplet corroboration}} \\
\rowcolor{lightblueblock}
SimVLA & SimVLA~\citep{luo2026simvla} & LIBERO & triplet & $20/20, 20/20, 0/20$ \\
\rowcolor{lightblueblock}
Executability & OpenPI-pi0.5 & LIBERO-10 & triplet & $19/20, 19/20, 1/20$ \\
\rowcolor{midblueblock}
Destination (hard) & OpenPI-pi0.5 & LIBERO-10 & triplet & $18/20, 19/20, 4/20$ \\
\addlinespace[2pt]
\rowcolor{lightgrayblock}
\multicolumn{5}{l}{\textit{Matched support comparisons}} \\
\rowcolor{midblueblock}
SSAC Order & OpenVLA-OFT & OFT-LIBERO & matched support & baseline $0.80, 1.00, 1.00$; support $1.00, 1.00, 1.00$ \\
\rowcolor{midblueblock}
Bowl-on-Ramekin & OpenVLA-OFT & LIBERO-Spatial & matched support & baseline $108/200, 84/200$; support $108/200, 140/200$ \\
\rowcolor{midblueblock}
Noise-band support & OpenVLA-OFT & LIBERO-Spatial & matched support & baseline $187/200$; support $191/200$ \\
\addlinespace[2pt]
\rowcolor{lightgrayblock}
\multicolumn{5}{l}{\textit{Native corroboration under local metrics}} \\
\rowcolor{deepblueblock}
LIBERO-10 TargetSwap & OpenPI-pi0.5 & LIBERO-10 & native corroboration & $0/20, 0/20, 0/20$; gap $+0.113$ \\
\rowcolor{deepblueblock}
Second-carrier corr. & OpenVLA-OFT & LIBERO-10 & native corroboration & $0/20, 0/20, 0/20$; gap $+0.005$ \\
\bottomrule
\end{tabular}
\end{adjustbox}
\end{table}

The matched-support block uses local comparator names from the supplementary evaluations. In particular, \emph{SSAC Order} denotes the order-support row from the SSAC comparator family on the OFT-LIBERO carrier: the evaluation reports a baseline-to-support clean/control/break improvement under the local OFT order protocol rather than a new protocol-identical BAS-VLA deployment row.

Table~\ref{tab:boundary-negative} collects selected boundary cases in three groups: harder semantic-breaking families, stronger preserving-side stress settings, and cross-benchmark mismatch cases. Some rows retain the triplet structure used in the main semantic-breaking evaluations, while others are reported in suite-native formats such as changed-only stress testing or clean-only evaluation. For the non-triplet rows, the Result column states the comparison explicitly rather than reusing triplet shorthand.
\begin{table}[tbp]
\centering
\small
\setlength{\tabcolsep}{8pt}
\caption{Selected boundary and negative cases across additional benchmarks and protocols. Triplet rows report clean, control, and break directly; non-triplet rows use their suite-native comparison format.}
\label{tab:boundary-negative}
\begin{adjustbox}{width=\textwidth}
\begin{tabular}{lllll}
\toprule
\textbf{Slice} & \textbf{Carrier} & \textbf{Suite} & \textbf{Protocol} & \textbf{Result} \\
\midrule
\rowcolor{lightblueblock}
\multicolumn{5}{l}{\textit{Semantic-breaking boundary cases}} \\
\rowcolor{lightblueblock}
Executability change & OpenPI-pi0.5 & LIBERO-10 & triplet & $0/20, 0/20, 1/20$ \\
\rowcolor{lightblueblock}
Spatial relation (hard) & OpenPI-pi0.5 & LIBERO-10 & triplet & $17/20, 18/20, 9/20$ \\
\rowcolor{lightblueblock}
Destination change (hard) & SimVLA~\citep{luo2026simvla} & LIBERO & triplet & $18/20, 17/20, 6/20$ \\
\addlinespace[2pt]
\rowcolor{lightgrayblock}
\multicolumn{5}{l}{\textit{Preserving-side stress settings}} \\
\rowcolor{midblueblock}
Clutter+noise stress & OpenVLA-OFT & LIBERO-Spatial & changed only & baseline 82.5\%; stress 70.0\%; BAS-VLA 74.5\% \\
\rowcolor{midblueblock}
Viewpoint/layout stress & OpenVLA-OFT & LIBERO-Spatial & changed only & baseline 78.0\%; BAS-VLA 79.5\% \\
\addlinespace[2pt]
\rowcolor{lightgrayblock}
\multicolumn{5}{l}{\textit{Cross-benchmark mismatch}} \\
\rowcolor{lightbeigeblock}
MetaWorld Door-Lock & OpenPI-pi0.5 & MetaWorld & triplet & $17/30, 16/30, 17/30$ \\
\bottomrule
\end{tabular}
\end{adjustbox}
\end{table}

Table~\ref{tab:action-space-support} reports a focused action-space view of the BAS-VLA Full mechanism variant on the preserving anchor and the evaluated support families under a matched OpenVLA-OFT protocol. Unlike the main semantic-breaking tables, which are organized around task-level success, this table isolates whether the action-space diagnostics follow the expected sign structure for semantic separation. Break-sep. rate denotes the fraction of episodes in which the break rollout is farther from the clean reference than the control rollout, so higher values indicate clearer separation. Clean-reference gap denotes the average difference $\mathrm{dist}(\mathrm{break}, \mathrm{clean})-\mathrm{dist}(\mathrm{control}, \mathrm{clean})$. Initial-chunk separation and executed-prefix separation denote the analogous break-minus-control differences computed on the first action chunk and on the executed comparison prefix, respectively. Under this interpretation, the preserving anchor should remain near zero or negative in the three distance-difference columns, whereas evaluated breaking families should yield positive values.
\begin{table*}[t]
\centering
\footnotesize
\setlength{\tabcolsep}{4pt}
\caption{Focused action-space diagnostics on the preserving anchor and evaluated support families under BAS-VLA Full. All rows use the matched OpenVLA-OFT ext6 protocol with five trials per family and a 24-step comparison prefix. Larger positive values indicate clearer semantic separation.}
\label{tab:action-space-support}
\begin{tabularx}{\textwidth}{>{\raggedright\arraybackslash}X>{\raggedright\arraybackslash}X>{\centering\arraybackslash}X>{\centering\arraybackslash}X>{\centering\arraybackslash}X>{\centering\arraybackslash}X}
\toprule
\textbf{Block} & \textbf{Family} & \makecell{\textbf{Break-sep.}\\\textbf{rate} $\uparrow$} & \makecell{\textbf{Clean-ref.}\\\textbf{gap} $\uparrow$} & \makecell{\textbf{Initial-chunk}\\\textbf{sep.} $\uparrow$} & \makecell{\textbf{Executed-prefix}\\\textbf{sep.} $\uparrow$} \\
\midrule
Preserving anchor & Semantic-preserving control & 20.0 & $-0.019$ & $-0.019$ & $-0.017$ \\
\addlinespace[2pt]
\rowcolor{lightgrayblock}
\multicolumn{6}{l}{\textit{Retained support families}} \\
Retained support & Spatial relation & 100.0 & $+0.055$ & $+0.055$ & $+0.079$ \\
Retained support & Target object & 60.0 & $+0.001$ & $+0.001$ & $+0.002$ \\
Retained support & Destination & 60.0 & $+0.011$ & $+0.011$ & $+0.008$ \\
\bottomrule
\end{tabularx}
\end{table*}

The reported signs are consistent with this interpretation: the preserving anchor remains negative, spatial relation shows the clearest positive separation, and the target-object and destination rows remain positive but smaller in magnitude.

\subsection{Uncertainty and Trial-Level Stability Notes}
Table~\ref{tab:uncertainty} complements exact rollout counts with Wilson 95\% intervals. The style-shift intervals support the preserving improvement, while the Milk-Swap intervals show high clean/control retention and low old-task success under semantic change. These intervals describe uncertainty in aggregate binary outcomes; repeated trials and the matched evaluation protocol provide the accompanying experimental context.

\FloatBarrier

\subsection{Comparison to Alternative Calibration Strategies}
Table~\ref{tab:internal-lineage} compares BAS-VLA to several related calibration strategies and deployment variants under the local settings used to probe each mechanism. The left block summarizes preserving-side strategies, ranging from generic appearance consensus to more targeted auxiliaries such as Peripheral Anomaly-Triggered Robust Action Rerouting (PARR) and Phase-Scoped Scene-Style Consistency (PSSC), while the right block summarizes breaking-side or deployment-level strategies such as the semantic-margin-only and instruction-margin calibrators. Arrow notation denotes before/after percentage changes for the corresponding strategy; it is not a clean/control/break tuple.

\begin{table}[tbp]
\centering
\footnotesize
\setlength{\tabcolsep}{3pt}
\providecommand{\altGeneric}{\raisebox{-0.30ex}{\textcolor[RGB]{225,120,120}{\Large$\bullet$}}}
\providecommand{\altTargeted}{\raisebox{-0.30ex}{\textcolor[RGB]{117,177,126}{\Large$\bullet$}}}
\providecommand{\altDeploy}{\raisebox{-0.30ex}{\textcolor[RGB]{156,126,204}{\Large$\bullet$}}}
\caption{Comparison to alternative calibration strategies for BAS-VLA. The left table reports preserving-side strategies and preserving-oriented deployments, and the right table reports breaking-side or deployment-level strategies. Arrow notation denotes before/after percentage changes rather than clean/control/break tuples.}
\vspace{0.15cm}
\label{tab:internal-lineage}
\begin{minipage}[t]{0.49\textwidth}
\centering
\textbf{Preserving-side strategies}\par\vspace{4pt}
\scriptsize
\begin{adjustbox}{width=\linewidth}
\begin{tabular}{>{\raggedright\arraybackslash}p{0.23\linewidth}>{\centering\arraybackslash}p{0.13\linewidth}>{\raggedright\arraybackslash}p{0.16\linewidth}>{\raggedright\arraybackslash}p{0.40\linewidth}}
\toprule
\textbf{Variant} & \textbf{Type} & \textbf{Slice} & \textbf{Result} \\
\midrule
\rowcolor{lightblueblock}
Appearance consensus & \altGeneric & illumi\-nation & \makecell[l]{$0.70 \rightarrow 0.82$;\\ $0.50 \rightarrow 0.74$} \\
\rowcolor{lightblueblock}
PARR & \altTargeted & clutter/\allowbreak noise & \makecell[l]{$46/50 \rightarrow 47/50$;\\ $46/50 \rightarrow 48/50$;\\ $165/200 \rightarrow 174/200$} \\
\rowcolor{lightblueblock}
PSSC & \altTargeted & style shift & \makecell[l]{$95.0 \rightarrow 100.0$;\\ $84/200 \rightarrow 126/200$} \\
\rowcolor{midblueblock}
Style-aux only & \altTargeted & Bowl-on-Ramekin style & \makecell[l]{$84/200 \rightarrow 132/200$;\\ clean held at $107/200$} \\
\rowcolor{deepblueblock}
\textbf{BAS-VLA (style-aux)} & \altDeploy & Bowl-on-Ramekin style & \makecell[l]{$84/200 \rightarrow 140/200$;\\ clean held at $108/200$} \\
\bottomrule
\end{tabular}
\end{adjustbox}
\end{minipage}\hfill
\begin{minipage}[t]{0.49\textwidth}
\centering
\textbf{Breaking-side strategies}\par\vspace{4pt}
\scriptsize
\begin{adjustbox}{width=\linewidth}
\begin{tabular}{>{\raggedright\arraybackslash}p{0.20\linewidth}>{\centering\arraybackslash}p{0.13\linewidth}>{\raggedright\arraybackslash}p{0.14\linewidth}>{\raggedright\arraybackslash}p{0.41\linewidth}}
\toprule
\textbf{Variant} & \textbf{Type} & \textbf{Slice} & \textbf{Result} \\
\midrule
\rowcolor{lightblueblock}
Instruction-margin & \altGeneric & positive rate & $0.50 \rightarrow 0.85$ \\
\rowcolor{lightblueblock}
Semantic-margin only & \altTargeted & Milk-Swap & \makecell[l]{$196/200, 195/200,$\\ $16/200$; gap $89.5$} \\
\rowcolor{lightblueblock}
\textbf{BAS-VLA (core)} & \altDeploy & Milk-Swap & \makecell[l]{$196/200, 195/200,$\\ $0/200$; gap $97.5$} \\
\rowcolor{lightblueblock}
\textbf{BAS-VLA (core)} & \altDeploy & OJ-Swap & \makecell[l]{$197/200, 193/200,$\\ $0/200$; gap $96.5$} \\
\rowcolor{deepblueblock}
\textbf{BAS-VLA (full)} & \altDeploy & full deploy. & \makecell[l]{native $20/20$;\\ full $20/20$;\\ matched break row\\ weaker than core} \\
\bottomrule
\end{tabular}
\end{adjustbox}
\end{minipage}
\par\vspace{4pt}
{\scriptsize \altGeneric\ generic comparator \hspace{1.2em} \altTargeted\ targeted mechanism variant \hspace{1.2em} \altDeploy\ BAS-VLA deployment variant}
\end{table}

These strategies are reported as mechanism probes rather than as direct substitutes for the final method. Generic appearance consensus serves as a baseline appearance-averaging strategy for illumination-change settings. Peripheral Anomaly-Triggered Robust Action Rerouting (PARR) is an inference-time dual-branch rerouting strategy that compares actions from the original observation and a periphery-suppressed observation, then switches only when both the periphery anomaly score and the branch disagreement are high. Phase-Scoped Scene-Style Consistency (PSSC) is an early-phase consistency gate that activates only under a style-specific cue and only during the first few replanning steps. On the breaking side, the semantic-margin-only row isolates the online semantic-calibration mechanism without the preserving auxiliary, while the instruction-margin row moves the intervention to an offline instruction-level calibration stage rather than the online action-calibration stage used by BAS-VLA. The BAS-VLA core rows provide matched Milk-Swap and OJ-Swap references under the main breaking protocol, and the BAS-VLA (full) row reports the effect of applying the full composition monolithically instead of the selective deployment adopted in the main text.

\begin{figure}[t]
\centering
\begin{subfigure}[t]{0.32\textwidth}
\centering
\includegraphics[width=\linewidth]{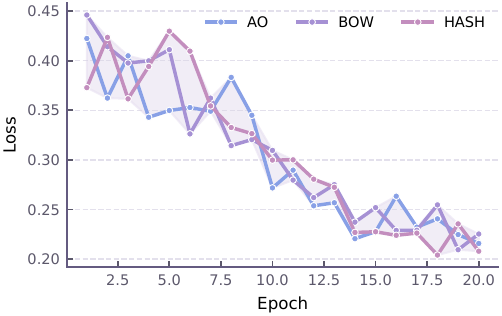}
\caption{Train total.}
\end{subfigure}\hfill
\begin{subfigure}[t]{0.32\textwidth}
\centering
\includegraphics[width=\linewidth]{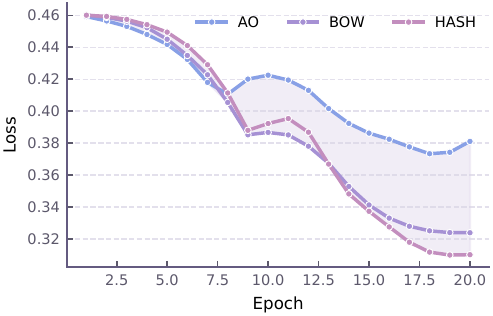}
\caption{Val total.}
\end{subfigure}\hfill
\begin{subfigure}[t]{0.32\textwidth}
\centering
\includegraphics[width=\linewidth]{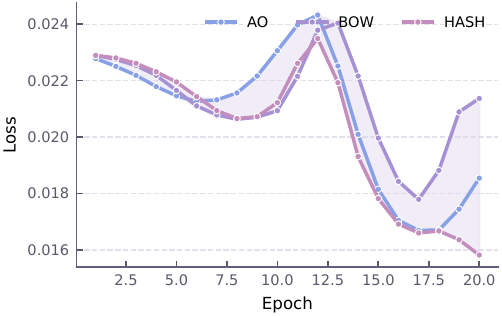}
\caption{Val consistency.}
\end{subfigure}
\caption{Metric-wise offline instruction comparator losses. Shaded bands show the per-epoch range across trainable variants.}
\label{fig:app-offline-loss-metricwise}
\end{figure}

\begin{figure}[t]
\centering
\begin{subfigure}[t]{0.32\textwidth}
\centering
\includegraphics[width=\linewidth]{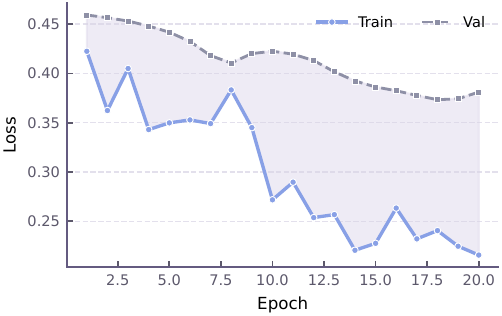}
\caption{AO.}
\end{subfigure}\hfill
\begin{subfigure}[t]{0.32\textwidth}
\centering
\includegraphics[width=\linewidth]{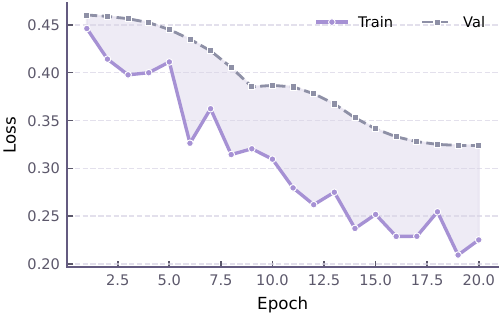}
\caption{BOW.}
\end{subfigure}\hfill
\begin{subfigure}[t]{0.32\textwidth}
\centering
\includegraphics[width=\linewidth]{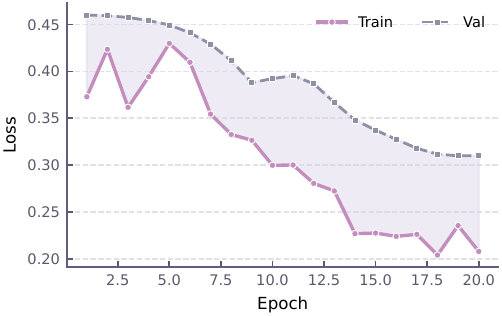}
\caption{HASH.}
\end{subfigure}
\caption{Per-variant train/validation trajectories for the offline instruction comparator family. Shaded regions mark train--validation gaps.}
\label{fig:app-offline-loss-variantwise}
\end{figure}

\begin{figure}[!htbp]
\centering
\begin{subfigure}[t]{0.32\textwidth}
\centering
\includegraphics[width=\linewidth]{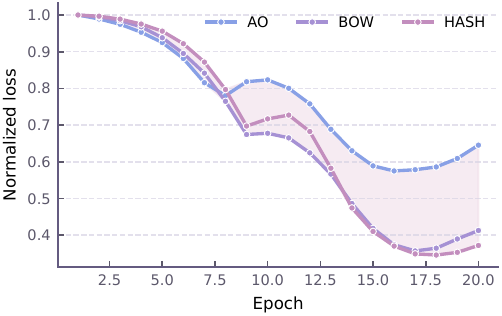}
\caption{Margin.}
\end{subfigure}\hfill
\begin{subfigure}[t]{0.32\textwidth}
\centering
\includegraphics[width=\linewidth]{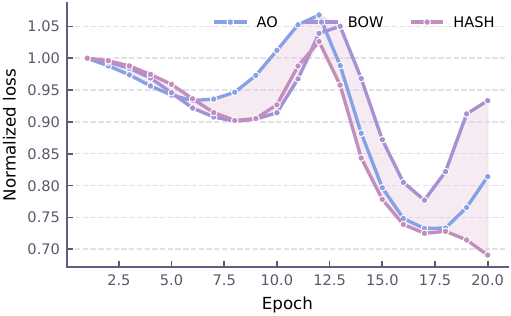}
\caption{Consistency.}
\end{subfigure}\hfill
\begin{subfigure}[t]{0.32\textwidth}
\centering
\includegraphics[width=\linewidth]{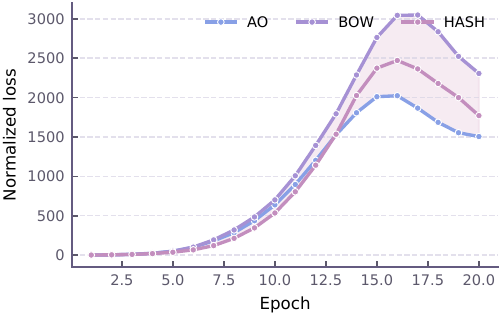}
\caption{Anchor.}
\end{subfigure}
\caption{Normalized loss profiles for the offline instruction comparator family. Shaded bands show per-epoch ranges across trainable variants.}
\label{fig:app-offline-loss-normalized}
\end{figure}

\begin{figure}[tbp]
\centering
\begin{subfigure}[t]{0.4\textwidth}
\centering
\includegraphics[width=\linewidth]{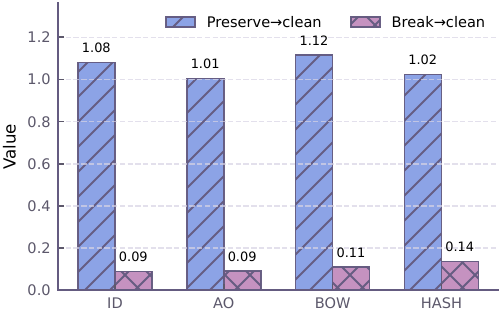}
\caption{Preserve vs.\ clean.}
\end{subfigure}\hfill
\begin{subfigure}[t]{0.4\textwidth}
\centering
\includegraphics[width=\linewidth]{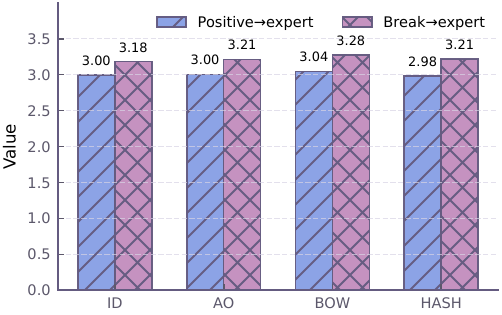}
\caption{Positive vs.\ expert.}
\end{subfigure}

\vspace{0.4em}

\begin{subfigure}[t]{0.4\textwidth}
\centering
\includegraphics[width=\linewidth]{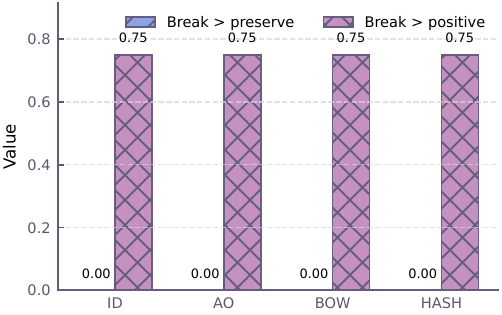}
\caption{Break separation.}
\end{subfigure}\hfill
\begin{subfigure}[t]{0.4\textwidth}
\centering
\includegraphics[width=\linewidth]{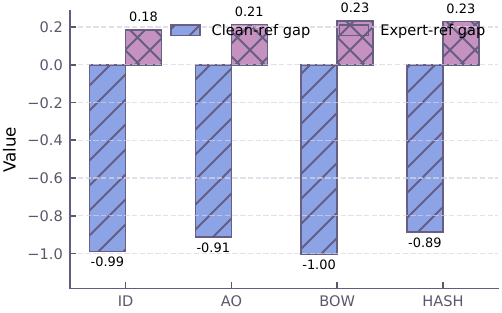}
\caption{Clean/expert gaps.}
\end{subfigure}
\caption{Aggregate action-gap diagnostics for the offline instruction comparator family.}
\label{fig:app-gap-summary}
\end{figure}

\begin{figure}[!b]
\centering
\begin{subfigure}[t]{0.32\textwidth}
\centering
\includegraphics[width=\linewidth]{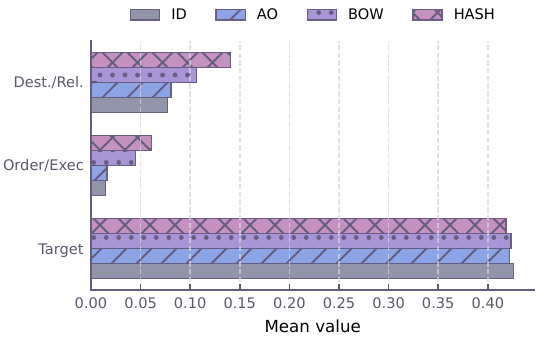}
\caption{Break-clean by family.}
\end{subfigure}\hfill
\begin{subfigure}[t]{0.32\textwidth}
\centering
\includegraphics[width=\linewidth]{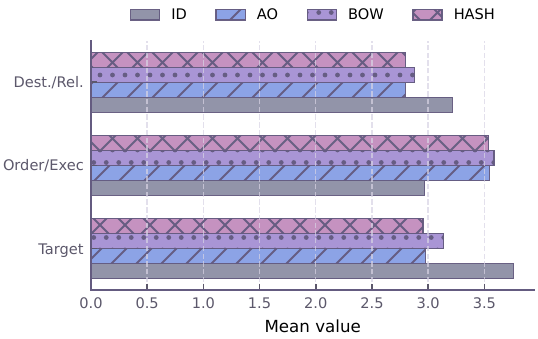}
\caption{Break-expert by family.}
\end{subfigure}\hfill
\begin{subfigure}[t]{0.32\textwidth}
\centering
\includegraphics[width=\linewidth]{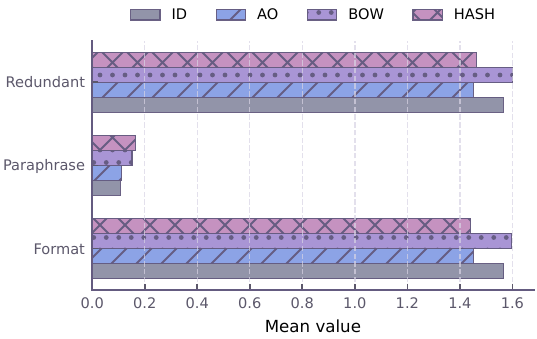}
\caption{Control invariance by family.}
\end{subfigure}
\caption{Family-wise action-distance diagnostics for the offline instruction comparator family.}
\label{fig:app-gap-familywise}
\end{figure}

Figures~\ref{fig:app-offline-loss-metricwise}--\ref{fig:app-gap-familywise} extend this subsection with process and diagnostic views for the offline instruction comparator family. The loss figures should be read as training-side mechanism evidence rather than as headline BAS-VLA training curves: they summarize how several comparator variants optimize their offline objectives. The action-gap figures then show what these variants do in action space, including how strongly break conditions separate from preserving references and how family-specific distances differ across target-object, ordering/executability, and destination/relation groups.

Figures~\ref{fig:app-offline-loss-metricwise}--\ref{fig:app-offline-loss-normalized} focus on optimization dynamics. Across the trainable variants, the total loss decreases over training, but the validation-side trajectories remain separated and the auxiliary objectives retain visible variant-dependent differences. The normalized views in Figure~\ref{fig:app-offline-loss-normalized} isolate trajectory shape from absolute scale and show that the trainable variants share a common early descent before diverging more clearly later in training.

Figures~\ref{fig:app-gap-summary} and~\ref{fig:app-gap-familywise} shift from training dynamics to action-space behavior. In the aggregate summaries, preserve-to-clean and positive-to-expert distances remain larger than the corresponding break-to-reference distances across the plotted variants, so the offline comparator family organizes these action relations differently from the main online calibration results. The family-wise breakdown further shows that break-to-clean distances are largest on target-object interventions, smaller on destination/relation slices, and smallest on ordering/executability slices, whereas the control-side distances are dominated by format and redundant-wording controls rather than by paraphrase.

\section{Limitations and Discussion}
\label{app:limitations-discussion}

The evidence supports a narrower claim than ``universal VLA robustness.'' BAS-VLA is best interpreted as a \emph{task-semantic calibration} method for cases in which task-relevant meaning truly changes. The strongest support comes from the semantic-breaking setting, where the method preserves clean/control competence while suppressing stale-task success under target change. Matched external comparisons support this interpretation, but they do not justify a claim that BAS-VLA dominates every inference-time strategy on every operational metric.

The preserving side is more limited. The auxiliary itself is defined by evidence-gated activation rather than by a family-specific switch, but the strongest preserving-side evidence currently comes from the validated style-shift slice rather than from a universal nuisance-stabilization regime. The full composition is not the preferred deployment under the present evidence. The appendix-level action-space diagnostics are likewise mechanism evidence rather than a replacement for task-level evaluation.

Supplementary benchmarks such as MetaWorld, CALVIN, ManiSkill2, and RoboCasa are best read as boundary or transfer evidence rather than as primary support for the central claim. Their role is to show where the same qualitative calibration logic remains visible and where it weakens under benchmark-specific interfaces or success criteria.

More broadly, the study remains concentrated on controlled intervention families and frozen-policy adaptation. Extending the trusted preserving families, handling more ambiguous semantic-breaking settings, and strengthening efficiency, uncertainty, and verification analyses remain open next steps.

\section{Appendix. Pseudocode and Formal Properties of BAS-VLA}

This appendix records an inference-time pseudocode summary together with several formal properties of the current BAS-VLA parameterization. The pseudocode fixes the deployed control flow and notation used throughout the paper, while the propositions below make explicit the local action-space implications of the breaking objective, the residual calibration form, the preserving-side gate and weak-fusion rule, and the resulting deployment-mode relations.

\subsection{Inference-Time Deployment Pseudocode}
\label{app:deployment-pseudocode}

Table~\ref{alg:bas-vla-inference} summarizes inference-time deployment under the three modes introduced in Section~4: the breaking-centered default mode, the preserving mode, and the full composition retained for ablation. The pseudocode uses exactly the same symbols as the main text and makes the evidence-gated control flow explicit.

\begin{table}[t]
\centering
\small
\caption{Inference-time deployment of BAS-VLA. The preserving-side gate uses the four factors from Eq.~\eqref{eq:pres-gate}; the concrete probe instantiation shown here uses text-conditioned grounding, mask refinement, and deterministic background attenuation.}
\label{alg:bas-vla-inference}
\begin{adjustbox}{width=\textwidth}
\begin{tabular}{@{}>{\raggedleft\arraybackslash}p{0.05\textwidth}>{\raggedright\arraybackslash}p{0.91\textwidth}@{}}
\toprule
\multicolumn{2}{@{}l@{}}{\textbf{Algorithm 1: Inference-time deployment of BAS-VLA}} \\
\midrule
\multicolumn{2}{@{}p{0.96\textwidth}@{}}{\textbf{Input:} observation $o_t$, instruction $x$, rollout context $h_t$, step index $t$, deployment mode $\mathrm{mode}\in\{\textsc{default},\textsc{pres},\textsc{full}\}$, frozen base policy $\pi_0$, frozen instruction feature extractor $z(\cdot)$, frozen visual feature maps $f_{\mathrm{vis}}^{\mathrm{mid}}, f_{\mathrm{vis}}^{\mathrm{late}}$, residual calibrator $r_\theta$, style-suppression map $\phi_{\mathrm{sty}}$, gate factors $\tau_{\mathrm{phase}}, g_{\mathrm{vis}}, g_{\mathrm{sem}}, g_{\mathrm{act}}$, scales $\kappa_{\mathrm{vis}}, \kappa_{\mathrm{sem}}$, residual scale $\lambda$, and fusion weight $\alpha$.} \\
\multicolumn{2}{@{}p{0.96\textwidth}@{}}{\textbf{Output:} deployed BAS-VLA action $a_t^{\mathrm{deploy}}$.} \\
\midrule
1 & Compute the frozen base action $a_t^{(0)} \leftarrow \pi_0(o_t, x, h_t)$. \\
2 & \textbf{if} $\mathrm{mode}=\textsc{pres}$ \textbf{then} \\
3 & \hspace*{1.5em}Ground task-relevant entities: $B_t \leftarrow \mathrm{GroundDINO}(o_t,x)$ for the robot arm, gripper, target object, and receptacle cues implied by $x$. \\
4 & \hspace*{1.5em}Refine a merged foreground mask $M_t \leftarrow \mathrm{SAM2}(o_t,B_t)$ and construct $\tilde{o}_t^{\mathrm{sty}} \leftarrow M_t \odot o_t + (1-M_t)\odot \psi_{\mathrm{bg}}(o_t)$ with deterministic background attenuation $\psi_{\mathrm{bg}}$. \\
5 & \hspace*{1.5em}Extract frozen mid/late visual features $v_t^{\mathrm{mid}}, \tilde{v}_t^{\mathrm{mid}}, v_t^{\mathrm{late}}, \tilde{v}_t^{\mathrm{late}}$ from the original and probe views. \\
6 & \hspace*{1.5em}Query the frozen policy on the probe view: $a_t^{\mathrm{sty}} \leftarrow \pi_0(\tilde{o}_t^{\mathrm{sty}}, x, h_t)$. \\
7 & \hspace*{1.5em}Compute $g_{\mathrm{vis}} \leftarrow \operatorname{clip}(\|v_t^{\mathrm{mid}}-\tilde{v}_t^{\mathrm{mid}}\|_2/\kappa_{\mathrm{vis}},0,1)$ and $g_{\mathrm{sem}} \leftarrow \exp(-\|v_t^{\mathrm{late}}-\tilde{v}_t^{\mathrm{late}}\|_2^2/\kappa_{\mathrm{sem}}^2)$. \\
8 & \hspace*{1.5em}Compute $\tau_t \leftarrow \tau_{\mathrm{phase}}(t,h_t)\, g_{\mathrm{vis}}\, g_{\mathrm{sem}}\, g_{\mathrm{act}}(a_t^{(0)},a_t^{\mathrm{sty}})$. \\
9 & \hspace*{1.5em}\textbf{return} $a_t^{\mathrm{deploy}} \leftarrow (1-\alpha\tau_t)a_t^{(0)} + \alpha\tau_t a_t^{\mathrm{sty}}$ \\
10 & \textbf{end if} \\
11 & Compute the frozen instruction feature $z_t \leftarrow z(x)$. \\
12 & Compute the breaking-centered calibrated action $a_t^{\mathrm{br}} \leftarrow a_t^{(0)} + \lambda\, r_\theta(a_t^{(0)}, z_t)$. \\
13 & \textbf{if} $\mathrm{mode}=\textsc{default}$ \textbf{then return} $a_t^{\mathrm{deploy}} \leftarrow a_t^{\mathrm{br}}$ \\
14 & Ground task-relevant entities, refine the foreground mask, and construct $\tilde{o}_t^{\mathrm{sty}}$ through the same $\phi_{\mathrm{sty}}$ pipeline; then query $a_t^{\mathrm{sty}} \leftarrow \pi_0(\tilde{o}_t^{\mathrm{sty}}, x, h_t)$. \\
15 & Extract frozen mid/late visual features from $o_t$ and $\tilde{o}_t^{\mathrm{sty}}$, compute $g_{\mathrm{vis}}$ and $g_{\mathrm{sem}}$, and then compute $\tau_t \leftarrow \tau_{\mathrm{phase}}(t,h_t)\, g_{\mathrm{vis}}\, g_{\mathrm{sem}}\, g_{\mathrm{act}}(a_t^{(0)},a_t^{\mathrm{sty}})$. \\
16 & \textbf{return} $a_t^{\mathrm{deploy}} \leftarrow (1-\alpha\tau_t)a_t^{\mathrm{br}} + \alpha\tau_t a_t^{\mathrm{sty}}$ \\
\bottomrule
\end{tabular}
\end{adjustbox}
\end{table}

\paragraph{Concrete probe instantiation note.}
The grounded-mask construction above is included as a concrete probe instantiation rather than as a new learning component. It enforces the same modeling intent used in the main text: preserve foreground entities that are semantically tied to the task, and attenuate only the nuisance-heavy complement region. The grounding and segmentation operators therefore define \emph{where} the probe may modify the observation, while the deterministic background attenuator $\psi_{\mathrm{bg}}$ defines \emph{how} the complement is weakened.

\subsection{Formal Action-Space Properties of the Breaking Objective}
\label{app:breaking-properties}

The breaking-centered objective is built around three terms: imitation on the clean and semantics-preserving control conditions, contraction of clean and control actions toward one another, and a margin-style separation term for semantic-breaking variants. The separation term is
\[
\mathcal{L}_{\mathrm{sep}}
=
\max\!\Big(
0,\;
m
- \|a^{\mathrm{br}}(x^b)-a^{\mathrm{br}}(x^c)\|_2
+ \|a^{\mathrm{br}}(x^u)-a^{\mathrm{br}}(x^c)\|_2
\Big).
\]

\paragraph{Proposition 1 (Margin implication).}
Assume $\mathcal{L}_{\mathrm{sep}} = 0$ for a triplet $(x^c,x^u,x^b)$. Then
\[
\|a^{\mathrm{br}}(x^b)-a^{\mathrm{br}}(x^c)\|_2
\ge
\|a^{\mathrm{br}}(x^u)-a^{\mathrm{br}}(x^c)\|_2 + m.
\]

\paragraph{Proof.}
By definition of the hinge loss, $\mathcal{L}_{\mathrm{sep}} = 0$ implies that the hinge argument is non-positive:
\[
m
- \|a^{\mathrm{br}}(x^b)-a^{\mathrm{br}}(x^c)\|_2
+ \|a^{\mathrm{br}}(x^u)-a^{\mathrm{br}}(x^c)\|_2
\le 0.
\]
Rearranging terms yields
\[
\|a^{\mathrm{br}}(x^b)-a^{\mathrm{br}}(x^c)\|_2
\ge
\|a^{\mathrm{br}}(x^u)-a^{\mathrm{br}}(x^c)\|_2 + m,
\]
which is the desired inequality. \hfill$\square$

This proposition states the basic local implication of the triplet loss: the semantic-breaking action must be separated from the clean action by at least margin $m$ beyond the clean-control discrepancy. The statement is local to the optimized action representation and does not by itself imply rollout-level correctness.

\subsection{Residual Drift Bounds Around the Frozen Base Policy}
\label{app:residual-bounds}

The breaking-centered core uses residual calibration,
\[
a_t^{\mathrm{br}} = a_t^{(0)} + \lambda\, r_{\theta}\!\left(a_t^{(0)}, z(x)\right),
\]
where $a_t^{(0)}$ is the frozen base-policy action and $r_\theta$ is the learned residual.

\paragraph{Proposition 2 (Residual drift bound).}
Suppose $\|r_\theta(a,z)\|_2 \le \varepsilon$ on a region of interest. Then the calibrated action satisfies
\[
\|a_t^{\mathrm{br}} - a_t^{(0)}\|_2 \le \lambda \varepsilon.
\]

\paragraph{Proof.}
Starting from the residual parameterization,
\[
\|a_t^{\mathrm{br}} - a_t^{(0)}\|_2
=
\|\lambda r_\theta(a_t^{(0)}, z(x))\|_2
=
\lambda \|r_\theta(a_t^{(0)}, z(x))\|_2
\le
\lambda \varepsilon.
\]
This proves the claim. \hfill$\square$

The bound shows that residual calibration stays within a radius controlled jointly by the residual magnitude and the scale parameter $\lambda$. An immediate consequence is that whenever $r_\theta(a_t^{(0)}, z(x)) = 0$ on a subset of states, BAS-VLA reduces exactly to the frozen base policy on that subset.

\subsection{Gate and Stability Properties of the Preserving Auxiliary}
\label{app:preserving-properties}

The preserving auxiliary uses the evidence gate
\[
\tau_t
=
\tau_{\mathrm{phase}}(t,h_t)\,
g_{\mathrm{vis}}(v_t^{\mathrm{mid}},\tilde{v}_t^{\mathrm{mid}})\,
g_{\mathrm{sem}}(v_t^{\mathrm{late}},\tilde{v}_t^{\mathrm{late}},x)\,
g_{\mathrm{act}}(a_t^{(0)},a_t^{\mathrm{sty}}),
\]
with each factor in $[0,1]$. In the current parameterization,
\[
g_{\mathrm{vis}}(v,\tilde{v})
=
\operatorname{clip}\!\left(\frac{\|v-\tilde{v}\|_2}{\kappa_{\mathrm{vis}}},\,0,\,1\right),
\qquad
g_{\mathrm{sem}}(v,\tilde{v},x)
=
\exp\!\left(-\frac{\|v-\tilde{v}\|_2^2}{\kappa_{\mathrm{sem}}^2}\right),
\]
where the first term is evaluated on a frozen mid-level visual representation and the second on a frozen later semantic visual representation from the same backbone. Under the concrete probe instantiation above, these features are extracted from the original image and from the grounded-mask probe image produced by deterministic background attenuation outside the retained foreground mask. As in the main text, the subscript ``sty'' denotes generic observation-side visual variation rather than only the validated style-shift slice. The preserving-only weak-fusion rule is
\[
a_t^{\mathrm{pres}}
=
(1-\alpha\tau_t)a_t^{(0)}+\alpha\tau_t a_t^{\mathrm{sty}},
\qquad \alpha\in(0,1).
\]
The full composition retained for ablation applies the same gate and fusion weight on top of the breaking-centered action,
\[
a_t^{\mathrm{full}}
=
(1-\alpha\tau_t)a_t^{\mathrm{br}}+\alpha\tau_t a_t^{\mathrm{sty}}.
\]

\paragraph{Proposition 3 (Gate range and identity cases).}
If $\tau_{\mathrm{phase}}, g_{\mathrm{vis}}, g_{\mathrm{sem}}, g_{\mathrm{act}} \in [0,1]$, then $\tau_t \in [0,1]$. Moreover,
\[
\tau_t = 0 \Longrightarrow a_t^{\mathrm{pres}} = a_t^{(0)}
\quad\text{and}\quad
a_t^{\mathrm{full}} = a_t^{\mathrm{br}}.
\]

\paragraph{Proof.}
The product of numbers in $[0,1]$ remains in $[0,1]$, hence $\tau_t \in [0,1]$. If $\tau_t=0$, then
\[
a_t^{\mathrm{pres}}
=
(1-\alpha\cdot 0)a_t^{(0)}+\alpha\cdot 0\,a_t^{\mathrm{sty}}
=
a_t^{(0)}.
\]
Applying the same substitution to the full-composition rule yields
\[
a_t^{\mathrm{full}}
=(1-\alpha\cdot 0)a_t^{\mathrm{br}}+\alpha\cdot 0\,a_t^{\mathrm{sty}}
=a_t^{\mathrm{br}}.
\]
\hfill$\square$

\paragraph{Proposition 4 (Gate-modulated drift identities).}
For every step,
\[
a_t^{\mathrm{pres}} - a_t^{(0)}
=
\alpha\tau_t \big(a_t^{\mathrm{sty}} - a_t^{(0)}\big),
\]
and therefore
\[
\|a_t^{\mathrm{pres}} - a_t^{(0)}\|_2
=
\alpha\tau_t \|a_t^{\mathrm{sty}} - a_t^{(0)}\|_2.
\]
In the full-composition mode,
\[
a_t^{\mathrm{full}} - a_t^{\mathrm{br}}
=
\alpha\tau_t \big(a_t^{\mathrm{sty}} - a_t^{\mathrm{br}}\big),
\]
and therefore
\[
\|a_t^{\mathrm{full}} - a_t^{\mathrm{br}}\|_2
=
\alpha\tau_t \|a_t^{\mathrm{sty}} - a_t^{\mathrm{br}}\|_2.
\]

\paragraph{Proof.}
Subtracting $a_t^{(0)}$ from the preserving-only weak-fusion equation gives
\[
a_t^{\mathrm{pres}} - a_t^{(0)}
=
=
\big((1-\alpha\tau_t)-1\big)a_t^{(0)}+\alpha\tau_t a_t^{\mathrm{sty}}
=
\alpha\tau_t\big(a_t^{\mathrm{sty}} - a_t^{(0)}\big),
\]
and taking the $\ell_2$ norm yields
\[
\|a_t^{\mathrm{pres}} - a_t^{(0)}\|_2
=
\|\alpha\tau_t(a_t^{\mathrm{sty}} - a_t^{(0)})\|_2
=
\alpha\tau_t\|a_t^{\mathrm{sty}} - a_t^{(0)}\|_2,
\]
because $\alpha\tau_t\ge 0$ is scalar. The full-composition identities follow by the same algebra after replacing the anchor action $a_t^{(0)}$ with $a_t^{\mathrm{br}}$. \hfill$\square$

\noindent
Two immediate consequences are useful. First, on every active step,
\[
a_t^{\mathrm{pres}} - a_t^{\mathrm{sty}}
=
(1-\alpha\tau_t)\big(a_t^{(0)} - a_t^{\mathrm{sty}}\big),
\]
and therefore
\[
\|a_t^{\mathrm{pres}} - a_t^{\mathrm{sty}}\|_2
=
\left(1-\alpha\tau_t\right)\|a_t^{(0)} - a_t^{\mathrm{sty}}\|_2.
\]
Together, these identities show that the preserving auxiliary remains on the line segment between its anchor action and the probe action, with displacement scaled continuously by $\tau_t$. The anchor is $a_t^{(0)}$ in preserving-only deployment and $a_t^{\mathrm{br}}$ in the full composition.

\subsection{Deployment-Mode Complexity and Identity Relations}
\label{app:deployment-properties}

This work uses three deployment modes:
\[
a_t^{\mathrm{default}} = a_t^{\mathrm{br}},\qquad
a_t^{\mathrm{pres}} = (1-\alpha\tau_t)a_t^{(0)}+\alpha\tau_t a_t^{\mathrm{sty}},\qquad
a_t^{\mathrm{full}} = (1-\alpha\tau_t)a_t^{\mathrm{br}}+\alpha\tau_t a_t^{\mathrm{sty}}.
\]

Let $C_{\pi_0}$ denote the cost of one frozen base-policy query, let $C_r$ denote the cost of one residual forward pass through $r_\theta$, and let $C_g$ collect the trigger/fusion overhead that does not invoke an additional base-policy query.

\paragraph{Proposition 5 (Coarse-grained per-mode overhead).}
Measured relative to a single frozen base-policy query, and under the convention that the mode-selection logic is absorbed into $C_g$, the additional inference cost of the three deployment modes satisfies
\[
\Delta C_{\mathrm{default}} = C_r + C_g,\qquad
\Delta C_{\mathrm{pres}} = C_{\pi_0}+C_g,\qquad
\Delta C_{\mathrm{full}} \ge C_r+C_g,
\]
where the preserving-side gate changes the fused action continuously but does not remove the need for the auxiliary probe query.

\paragraph{Proof.}
The default mode adds one residual pass and the mode-selection overhead on top of the frozen-policy output, hence $\Delta C_{\mathrm{default}} = C_r + C_g$. The preserving sidecar computes the probe action together with the gate and fusion overhead, hence $\Delta C_{\mathrm{pres}} = C_{\pi_0}+C_g$. The full mode contains the residual pass and the full-composition overhead and is therefore bounded below by $C_r+C_g$. \hfill$\square$

The current parameterization also contains several exact identity relations:
\[
r_\theta(\cdot,\cdot)\equiv 0
\Longrightarrow
a_t^{\mathrm{br}} = a_t^{(0)},
\]
\[
\tau_t = 0
\Longrightarrow
a_t^{\mathrm{pres}} = a_t^{(0)}
\quad\text{and}\quad
a_t^{\mathrm{full}} = a_t^{\mathrm{br}},
\]
\[
\alpha = 0
\Longrightarrow
a_t^{\mathrm{pres}} = a_t^{(0)}
\quad\text{and}\quad
a_t^{\mathrm{full}} = a_t^{\mathrm{br}}
\quad\text{even on triggered steps.}
\]
These identities clarify that BAS-VLA remains a calibration layer around the frozen policy and that exact recovery of the base action is contained in the parameterization.

\end{document}